%% file: accv2020cameraready.tex
\documentclass[runningheads]{llncs}
\input{math_commands.tex}

\usepackage{times}
\usepackage{epsfig}
\usepackage{microtype}
\usepackage{graphicx}
\usepackage{amsmath}
\usepackage{amssymb}
\usepackage{xcolor}
\usepackage[T1]{fontenc}
\usepackage{ownstyles}
\usepackage{booktabs} 
\usepackage{color}

\usepackage{hyperref}

\usepackage[normalem]{ulem}
\usepackage{bm}
\usepackage{dsfont}
\usepackage{diagbox}
\usepackage{array}
\usepackage{multirow}
\usepackage{nccmath}
\usepackage{caption}
\usepackage{subcaption}
\usepackage{float} 
\usepackage{makecell}
\usepackage{ulem}

\newcommand{\class}[1]{\ensuremath{\mathsf{#1}\xspace}}

\usepackage{ownstyles}

\newif\ifcomments%

\commentsfalse

\ifcomments%
\newcommand{\comments}[1]{#1}
\else
\newcommand{\comments}[1]{}
\fi

\newcommand{\eg}{e.g.\xspace}
\newcommand{\ie}{i.e.\xspace}

\newcommand{\subsec}[1]{\noindent{\textbf{#1~~}}}

\begin{document}
\pagestyle{headings}
\mainmatter

\def\ACCV20SubNumber{708}  

\title{Explaining image classifiers by removing input features using generative models} 
\titlerunning{Explaining an image classifier's decisions using generative models}
\author{Chirag Agarwal\inst{} \and
Anh Nguyen\inst{}}
%
\authorrunning{Chirag Agarwal and Anh Nguyen}
%
\institute{Auburn University, Auburn, AL 36849, USA
\email{\{chiragagarwall12,anh.ng8\}@gmail.com}
}

\maketitle

\begin{abstract}
	Perturbation-based explanation methods often measure the contribution of an input feature to an image classifier's outputs by heuristically removing it via \eg blurring, adding noise, or graying out, which often produce unrealistic, out-of-samples.
	Instead, we propose to integrate a generative inpainter into three representative attribution methods to remove an input feature.
	Our proposed change improved all three methods in (1) generating more plausible counterfactual samples under the true data distribution; 
	(2) being more accurate according to three metrics: object localization, deletion, and saliency metrics;
	and (3) being more robust to hyperparameter changes.
	Our findings were consistent across both ImageNet and Places365 datasets and two different pairs of classifiers and inpainters.
\end{abstract}

\section{Introduction}
Explaining a classifier's outputs given a certain input is increasingly important, especially for life-critical applications \cite{doshi2017towards,gunning2019darpa}.
A popular means for visually explaining an image classifier's decisions is an \emph{attribution map} \ie a heatmap that highlights the input pixels that are the evidence for and against the classification outputs \cite{montavon2018methods}.
To construct an attribution map, many methods approximate the attribution value of an input region by the classification probability change when that region is absent \ie removed from the image.
While removing an input feature to measure its attribution is a principle method (\ie ``intervention'' in causal reasoning \cite{hagmayer2007causal}), a key open question is: \textbf{How to remove?}

\begin{figure}[ht]
	\centering
	{
		\begin{flushleft}
			\hskip -0.1in
            \rotatebox{90}{\kern -10.25pc heatmaps\kern 3.5pc samples}
			\hskip -0.05in (a) Real + BB
			\hskip 0.07in (b) SP \cite{zeiler2014visualizing}
			\hskip 0.15in (c) SP-G
			\hskip 0.1in (d) LIME \cite{ribeiro2016should}
			\hskip 0.01in (e) LIME-G
			\hskip 0.15in (f) MP2
			\hskip 0.1in (g) MP2-G
		\end{flushleft}
	}
    \vskip -0.1in
	{
			\includegraphics[width=0.97\textwidth]{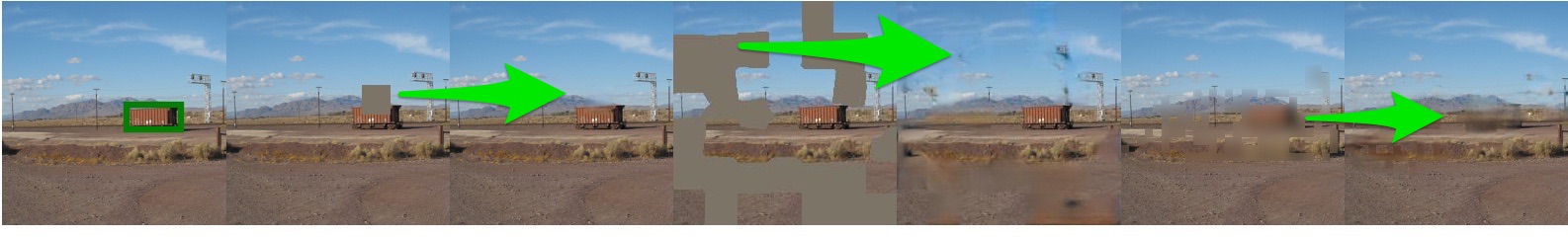}
	}
	{	
		\footnotesize
		\begin{flushleft}
            \vskip -0.1in
			\hskip -0.07in \class{freight~car}~0.832
			\hskip 0.1in 0.391
			\hskip 0.3in 0.840
			\hskip 0.4in 0.003
			\hskip 0.35in 0.898
			\hskip 0.4in 0.001
			\hskip 0.3in 0.001
		\end{flushleft}
	}
	{
		\begin{flushright}
			\includegraphics[width=0.99\textwidth]{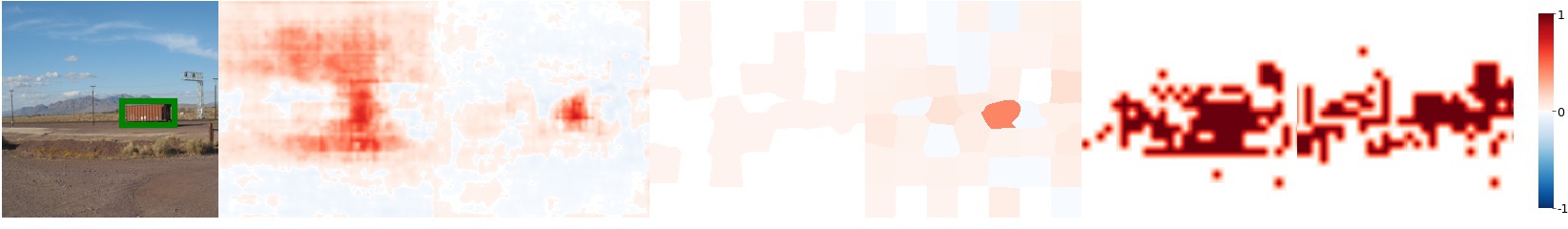}
		\end{flushright}
	}
	\vskip -0.1in
	\caption{
		Three attribution methods, SP \cite{zeiler2014visualizing}, LIME \cite{ribeiro2016should}, and MP2, often produce unrealistic, out-of-distribution perturbation samples.
		\textbf{Top row:} SP slides a $29\times29$ gray patch across the image (b).
		LIME grays out a set of random superpixels (d).
		MP2 blurs out the entire image (f).
		In contrast, a learned inpainter integrated into these methods produces realistic samples for the same perturbation masks, here, completing the \class{freight~car} (c), completing the background (e), and removing the car from the scene (g).
		Note that the \class{freight~car} class probability is reduced by 53\% (\ie from 0.832 to 0.391) when only a part of the car was occluded (b).
		However, it is reduced by $\sim$100\% down to 0.003 when the car is still present but the background is unnaturally masked out (d).
		Since the inpainted samples are more realistic, the probability drops are often less (c \& e) and substantial only when the object is removed completely (g).
		\textbf{Bottom row:} the inpainted samples yield heatmaps that, in overall, outperform the original methods on the object localization task Sec.~\ref{sec:faithful}. 
		Here, our heatmaps (SP-G, LIME-G, and MP2-G) are less noisy and more focused on the object. 
	}
	\vskip -0.1in
	\label{fig:teaser}
\end{figure}

State-of-the-art perturbation-based attribution methods implement the absence of an input feature by replacing it with (a) mean pixels \cite{zeiler2014visualizing,ribeiro2016should}; (b) random noise \cite{dabkowski2017real,lundberg2017unified}; or (c) blurred versions of the original content \cite{fong2017interpretable,fong2019understanding}.
However, these removal (\ie perturbation) techniques often produce unrealistic, out-of-distribution images (Fig.~\ref{fig:teaser}b,d) on which the classifiers were not trained.
Because classifiers are often easily fooled by unusual input patterns \cite{alcorn2019strike,nguyen2015deep,agarwal2019improving}, we hypothesize that such examples might yield heatmaps that are (1) unreliable \ie sensitive to hyperparameter settings \cite{bansal2020sam}; and (2) not faithful \cite{adebayo2018sanity}.

To combat these two issues, we propose to harness a state-of-the-art generative inpainting model (hereafter, an inpainter) to remove pixels from an input image and fill in with content that is plausible under the true data distribution.
We test our approach on three representative attribution methods of Sliding-Patch (SP) \cite{zeiler2014visualizing}, LIME \cite{ribeiro2016should}, and Meaningful-Perturbation (MP) \cite{fong2017interpretable} across two large-scale datasets of ImageNet \cite{russakovsky2015imagenet} and Places365 \cite{zhou2017places}.
For each dataset, we use a separate pair of pre-trained image classifiers and inpainters.
We chose SP, LIME, and MP because they are among the most commonly used and applicable to any classifier.
Our main findings include:\footnote{
All our codes are available at \href{https://github.com/anguyen8/generative-attribution-methods}{https://github.com/anguyen8/generative-attribution-methods}.}
\begin{enumerate}
    \item Inpainting is more effective than common techniques in removing discriminative features. That is, photos with the main object blurred or grayed out are still 3 times more recognizable by classifiers and more similar to the original photo (via MS-SSIM and LPIPS) than photos with objects removed via inpainting (Sec.~\ref{sec:discriminative}).
	\item Our results are the first to show that incorporating an inpainter improves perturbation-based attribution methods \ie producing (1) more plausible perturbation samples; (2) explanations that are similarly or more accurate on three common benchmarks---object localization, deletion, and saliency metrics (Sec.~\ref{sec:faithful}); and (3) explanations that are more robust to hyperparameter changes \ie the SAM metric \cite{bansal2020sam} (Sec.~\ref{sec:sensitivity}); 
    \item We propose MP2-G (Sec.~\ref{sec:MP}), a variant that is substantially more accurate, reliable, and having four hyperparameters fewer than the common MP \cite{fong2017interpretable}---a state-of-the-art approach which is the basis for many extensions
\cite{wagner2019interpretable,qi2019visualizing,carletti2018understanding,wang2018learning,uzunova2019interpretable}. 
    
\end{enumerate}

To the best of our knowledge, this is the first work that shows the effectiveness of generative models in improving the accuracy and reliability of explanation methods.

\section{Related work}
\label{sec:related_work}

Attribution methods can be grouped into two main classes: (1) white- and (2) black-box.

\noindent \textbf{White-box~} Given access to the network architecture and parameters, attribution maps can be constructed analytically from (a) the gradients of the output w.r.t. the input \cite{simonyan2013deep}, (b) the class activation map in fully-convolutional neural networks \cite{zhou2016learning}, (c) both the gradients and activations \cite{selvaraju2017grad}, or (d) the gradient times the input image \cite{shrikumar17learning}.
However, some gradient-based heatmaps can be too noisy to be human-interpretable \cite{smilkov2017smoothgrad}, and suffer from gradient saturation \cite{sundararajan2017axiomatic}.
To combat these issues, perturbation techniques were also utilized.
That is, to make a gradient-based heatmap more robust and smooth, a number of methods essentially average out the resultant heatmaps across a large set of perturbed inputs that are created via (a) adding random noise to the input \cite{fong2017interpretable,smilkov2017smoothgrad}, (b) blurring the input \cite{fong2017interpretable}, or (c) linearly interpolating between the input and a reference ``baseline'' image \cite{sundararajan2017axiomatic}.

\noindent \textbf{Black-box~}
Perturbation-based methods are important for use cases when only a black-box model is given (no network parameters). 
Black-box methods often remove (\ie perturb) an input region and take the resultant classification probability change to be the attribution value for that region.
While the idea is principle in causal reasoning, the physical interventions---taking an object out of a scene (revealing the content behind it) while keeping other factors unchanged---are impractical in most real-world applications.
The absence of an input region is often implemented by replacing it with (a) mean pixels \cite{zeiler2014visualizing,ribeiro2016should}; (b) random noise \cite{dabkowski2017real,lundberg2017unified}; or (c) blurred versions of the original content \cite{fong2017interpretable}.
However, these removal techniques often produce unrealistic, out-of-samples (Fig.~\ref{fig:teaser}), which raises huge concerns on the sensitivity and faithfulness of explanations.

\noindent \textbf{An open question} for existing perturbation-based attribution methods is: \emph{Do explanations become more robust and accurate if input features are removed via a strong, natural image prior?} 
Here, we systematically study this question across three representative attribution methods: two black-box methods that are perturbation-based (\ie SP and LIME) and one white-box method that relies on both perturbations and gradients (\ie MP).
These representative methods also perturb different types of input features: pixels (\ie MP), superpixels (\ie LIME); and  square patches (\ie SP).

The closest to our work is FIDO-CA \cite{chang2019explaining}, which extended MP and harnessed an image inpainter to synthesize counterfactual samples to explain classifiers' decisions.
However, FIDO-CA \cite{chang2019explaining} underperformed most baselines that do not use inpainters.
Inspired by \cite{chang2019explaining}, we propose a key change in optimization objectives (see details in Sec.~\ref{sec:results}) that enabled our approach to improve upon FIDO-CA by a large margin.
That is, for the first time, we show that incorporating an inpainter improves the accuracy and robustness of explanation methods.

\section{Methods}
\label{sec:methods}
\subsection{Datasets and Networks}
\label{sec:networks}
\subsec{Classifiers} Our experiments were conducted using two separate ResNet-50 image classifiers \cite{he2016deep} that were pre-trained on the 1000-class ImageNet 2012 \cite{russakovsky2015imagenet} and Places365 \cite{zhou2017places}, respectively.
The two models were officially released by the PyTorch model zoo \cite{torchvis88:online} and by the authors \cite{CSAILVis16:online}, respectively.

\subsec{Datasets}
We chose these two datasets because they are large, natural-image sets covering a wide range of images from object-centric (\ie ImageNet) to scenery (\ie Places365).
From the 50,000 ImageNet and 36,500 Places365 validation-set images, we randomly sampled a set of 2000 images correctly classified by their respective ResNet-50 models.
We used these two sets of images
in all experiments throughout the paper.

\subsec{Inpainters}
For each classifier, pre-trained either on ImageNet or Places365, we used a 
TensorFlow DeepFill-v1 model pre-trained by \cite{yu2018generative} on the same respective dataset.
DeepFill-v1 takes as input a color image and a binary mask, both at resolution $256\times 256$, and outputs an inpainted image of the same size.
In this work, we also tried DeepFill-v2 \cite{yu2018free}, a free-form inpainting model, but the overall results did not change significantly.
Apart from these two, to the best of our knowledge, there are no other publicly available generative inpainters for both ImageNet and Places365 datasets.
The DeepFill-v1 inpainter is practically feasible for attribution algorithms as it only takes 0.2s/image on one GPU (and 1.5s/image on CPUs) for inpainting a $512 \times 512$ image.

\subsection{Problem formulation}
Let $s:\sR^{D\times D \times 3} \to \sR$ be an image classifier that maps a square, color image $\vx$ of spatial dimension $D \times D$ onto a softmax probability of a target class.
An attribution map $\mA \in [-1,1]^{D\times D}$ associates each input pixel $\vx_i$ to a scalar $\mA_i \in [-1,1]$ which indicates how much $\vx_i$ contributes for or against the prediction score $s(\vx)$.
We describe below three methods for generating attribution maps together with our respective proposed variants (hereafter, G-methods) which harness a generative inpainter.

\subsection{Sliding-Patch (SP)}
\label{sec:method:SP}

\textbf{SP}
\cite{zeiler2014visualizing} proposed to slide a gray, occlusion patch across the image and record the probability changes as attribution values in corresponding locations in the heatmap.
That is, given a binary occlusion mask $m \in \{0, 1\}^{D\times D}$ (here, $1$'s inside the patch region and $0$'s otherwise) and a filler image
$\vf \in \sR^{D\times D\times3}$, a perturbed image $\bar{\vx} \in \sR^{D\times D\times3}$ (see Fig.~\ref{fig:teaser}b) is given by:
\begin{equation}
\bar{\vx} = \vx \odot (1 - m) + \vf \odot m
\label{eqn:infill}
\end{equation}

\noindent
where $\odot$ denotes the Hadamard product and $\vf$ is a zero image \ie a gray image\footnote{The ImageNet mean pixel is gray (0.485, 0.456, 0.406).} before input pre-processing.
For every pixel $\vx_i$, one can generate a perturbation sample $\bar{\vx}^i$ (\ie by setting the patch center at $\vx_i$) and compute the attribution value $\mA_i = s(\vx) - s(\bar{\vx}^i)$.
However, sliding the patch densely across the $224\times224$ input image is prohibitively slow.
Therefore, we chose a $29\times 29$ occlusion patch size with stride 3, which yields a smaller heatmap $\mA'$ of size $66\times 66$.
We bi-linearly upsampled $\mA'$ to the image size to create the full-res $\mA$.
See Fig.~\ref{fig:teaser}b for an example of SP heatmaps and perturbed images.

We implemented SP by converting a MATLAB implementation \cite{matlab2019occlusion} into PyTorch.
All of our individual experiments in this work were run on a single GTX 1080Ti GPU.

\subsec{SP-G}
Note that the stride, size, and color of a SP sliding patch are three hyperparameters that are often chosen heuristically, and varying them can change the final heatmaps radically \cite{bansal2020sam}.
To ameliorate the sensitivity to hyperparameter choices, we propose a variant called SP-G by only replacing the gray filler image of SP with the output image of an inpainter (described in Sec.~\ref{sec:networks}) \ie $\vf = G(m, \vx)$ while keeping the rest of SP the same (Fig.~\ref{fig:teaser}b~vs.~c; top row).
That is, at every location of the sliding window, SP-G queries the inpainter for content to fill in the window.
\subsection{Local Interpretable Model-Agnostic Explanations (LIME)}
\subsec{LIME} 
While SP occludes one square patch of the image at a time, LIME \cite{ribeiro2016should} occludes a random-shaped region.
The algorithm first segments the input image into $S$ non-overlapping superpixels \cite{achanta2012slic}.
Then, LIME generates a perturbed image $\bar{\vx}$ by graying out a random set of superpixels among $2^S$ possible sets.
That is, LIME follows Eq.~\ref{eqn:infill} where the pixel-wise mask $m$ is derived from a random superpixel mask $m' \in \{0,1\}^S$.
For each sample $\bar{\vx}^i$, we measure the output score $s(\bar{\vx}^i)$ and evenly distribute it among all occluded superpixels in $\bar{\vx}^i$. 
Each superpixel's attribution is then inversely weighted by the $L_2$ distance $\Vert \vx - \bar{\vx}^i \Vert$ via an exponential kernel and then averaged out across $N$ samples.
The resultant attribution $\va_k$ of a superpixel $k$ is finally assigned to all pixels in that group in the full-resolution heatmap $\mA$.

In practice, \cite{ribeiro2016should} iteratively optimized for $\{\va_k\}_S$ via LASSO for $1000$ steps to also maximize the number of zero attributions \ie encouraging simpler, sparse attribution maps.
We used the implementation provided by the authors of LIME \cite{lime2019code} and their default hyperparameters of $S = 50$ and $N = 1000$.

\subsec{LIME-G}
While avoiding the bias given by the SP square patch, LIME perturbation samples remain unrealistic.
Therefore, we propose LIME-G, a variant of LIME, by only changing the gray image $\vf$ to a synthesized image $G(m, \vx)$ as in SP-G while keeping the rest of LIME unchanged.

\subsection{Meaningful Perturbation (MP)}
\label{sec:MP}
\textbf{MP~~} As SP and LIME gray out patches and superpixels in the input image, they generate unrealistic counterfactual samples and produce coarse heatmaps.
To combat these issues, Fong et al.~\cite{fong2017interpretable} proposed the MP algorithm \ie learning a minimal, continuous mask $m \in [0,1]^{D\times D}$ that blurs out the input image in a way that would minimize the target-class probability.
That is, MP attempts to solve the following optimization problem:
\begin{equation}
m^* = \argmin_{m} \lambda \Vert m \Vert_1 + s( \bar{\vx})
\label{eq:mp_original}
\end{equation}
where $\bar{\vx}$ is given by Eq.~\ref{eqn:infill} but with $\vf = B_\sigma(\vx)$ \ie the input image blurred by a Gaussian kernel $B_\sigma(.)$ of radius $\sigma = 10$.
Note that, in MP, the attribution map $\mA$ is also the learned mask $m$.
However, solving Eq.~\ref{eq:mp_original} directly often yields heatmaps that are noisy and sensitive to hyperparameter changes \cite{bansal2020sam}.
Therefore, MP only learned a small $28\times28$ mask and upsampled it to the image size in every optimization step.
In addition, they also encouraged the mask to be smooth and robust to input changes by changing the objective function to the following:
\begin{equation}
m^* = \argmin_{m} \lambda_1 \Vert m \Vert_1 + \lambda_2 TV(m) + \E_{\tau \sim \mathcal{U}(0, 4)} \big[ s( \Phi(\bar{\vx}, \tau) ) \big]
\label{eq:mp_full}
\end{equation}
where $TV(m) = \sum_{i} \Vert \nabla m_i \Vert_{3}^{3}$ \ie a total-variation norm that acts as a smoothness prior over the mask.
The third term is the expectation over a batch of randomly jittered versions of the blurred image $\bar{\vx}$.
That is, $\Phi(.)$ is the jitter operator that translates an image $\bar{\vx}$ vertically or horizontally by $\tau$ pixels where $\tau$ is drawn from a discrete uniform distribution $\mathcal{U}(0,4)$.
We randomly initialized the mask from a continuous uniform distribution $\mathcal{U}(0, 1)$ and minimize the objective function in Eq.~\ref{eq:mp_full} via gradient descent for $300$ steps.
Our MP implementation was in PyTorch and followed all the hyperparameters as described in \cite{fong2017interpretable}.


\noindent \textbf{MP2~~} 
In the original formulation, MP is highly sensitive to changes in some of its hyperparameters \cite{bansal2020sam}.
In our preliminary experiments (data not shown), we found that integrating an inpainter into the existing unstable MP optimization did not yield more accurate heatmaps.
In addition, the $L_{1}$ and $TV$ terms (Eq.~\ref{eq:mp_full}) introduce strong biases that impede the contribution of the content generated by inpainters.



Therefore, we propose MP2, a more reliable and accurate variant by eliminating four hyperparameters from MP: the $L_{1}$ norm, $TV$ norm, the jitter operator and the stopping criterion of 300 optimization steps (Sec.~\ref{sec:faithful}).
That is, we still find a minimal mask (Eq.~\ref{eq:mp_original}) but by initializing it with all zeros and growing the number of 1's (\ie the blurred region) gradually.
Following JSMA \cite{papernot2016limitations}, in every iteration, we add 1's to two pixels that have the highest gradient norms.
We stop the mask optimization when the classification probability reaches random chance, \ie $0.001$ for ImageNet and $0.003$ for Places365.
As MP, we use the same Gaussian blur radius of 10 and the mask size of 28$\times$28.



\noindent\textbf{MP2-G~~} We integrate an inpainter $G$ to MP2 by only changing the filler image $\vf = B_\sigma(\vx)$ that is used in Eq.~\ref{eqn:infill} to an inpainted image \ie $\vf = G(m_{\text{b}}, \vx)$ where $m_{\text{b}} \in \{0,1\}^{D\times D}$ is the binary mask learned via MP2 optimization.


\section{Experiments and Results}
\label{sec:results}
\subsection{Inpainter failed to synthesize backgrounds given only foreground objects}
\label{sec:fido}
Chang et al. \cite{chang2019explaining} proposed to find a minimal set of input pixels that would keep the classification outputs unchanged even when the other pixels in the image are removed (\ie the ``preservation'' objective \cite{fong2017interpretable}) via an inpainter.
Their method, FIDO-CA, uses the same DeepFill-v1 inpainter as in our work; however, their ``preservation'' objective encourages the inpainter to predict the missing background pixels conditioned on the remaining foreground object---a task that DeepFill-v1 was \emph{not} trained to do and thus produced unrealistic samples as in \cite{chang2019explaining}.
In contrast, our MP2-G method harnesses the dual ``deletion'' objective \ie finding the smallest set of input pixels which when inpainted would minimize the target-class probability---which intuitively asks the inpainter to replace the main object with some content that is consistent with the background \ie the training objective of DeepFill-v1.

\begin{figure}[ht]
	\centering
	{	
		\begin{flushleft}
			\hskip 0.25in (a)
			\hskip 0.4in (b)
			\hskip 0.4in (c)
			\hskip 0.45in (d)
			\hskip 0.4in (e)
			\hskip 0.45in (f)
			\hskip 0.4in (g)
			\hskip 0.4in (h)
		\end{flushleft}
	}
	{	
		\begin{flushleft}
       		\vskip -0.1in
			\hskip 0.2in Real
			\hskip 0.3in Mask
			\hskip 0.1in Preserve \cite{chang2019explaining}
			\hskip 0.07in Delete
			\hskip 0.25in Real
			\hskip 0.33in Mask
			\hskip 0.1in Preserve \cite{chang2019explaining}
			\hskip 0.02in Delete
		\end{flushleft}
	}
    \vskip -0.1in
	\includegraphics[width=0.99\linewidth]{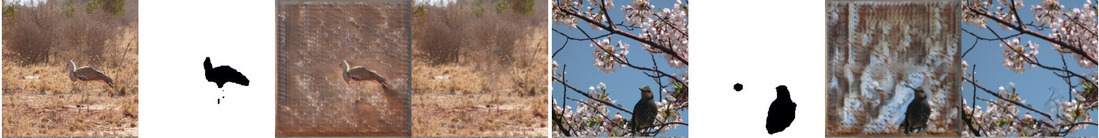}
	\caption{
        Using the DeepFill-v1 inpainter to fill in the background region (\ie ``preservation'' task \cite{chang2019explaining}) yields unrealistic images that contain features unnaturally pasted from the object (c, g).
        This key difference between the ``deletion'' (d, h) and ``preservation'' (c, g) objectives is further reflected in the evaluation results of MP2-G and FIDO-CA \cite{chang2019explaining} where the attribution maps generated using the latter consistently underperforms than MP2-G (Sec.~\ref{sec:faithful}).
		See Fig.~\ref{fig:FIDO_full} for more examples of the images.
	}
	\vskip -0.05in
	\label{fig:FIDO_cropped}
\end{figure}

\begin{figure}[ht]
	\centering
	{	
		\begin{flushleft}
			\hskip 0.9in (a) Real
			\hskip 0.2in (b) Blur
			\hskip 0.2in (c) Gray
			\hskip 0.15in (d) Inpaint
			\hskip 0.05in (e) Noise
		\end{flushleft}
	}
	\vskip -0.1in
	\includegraphics[width=0.65\columnwidth]{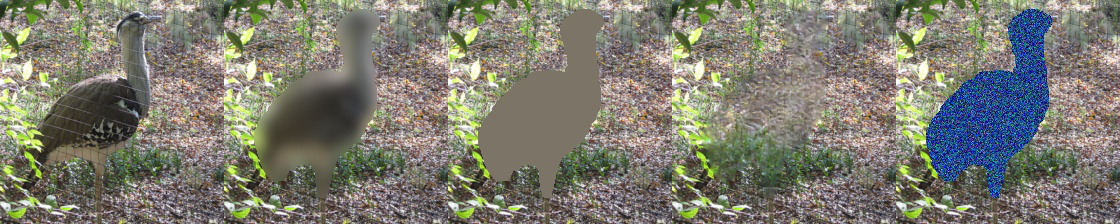}
	{	
		\begin{flushleft}
		    \vskip -0.1in
			\hskip 0.8in \class{bustard}~0.996
			\hskip 0.08in 0.020
			\hskip 0.3in 0.050
			\hskip 0.3in 0.001
			\hskip 0.3in 0.001
		\end{flushleft}
	}
    \vskip -0.1in
	\caption{
		The results of filling the object mask in a real image (a) via four different filling methods. 
		The shape of the bird is still visible even after blurring (b), graying out (c) or adding random noise (e) to the bird region.
		The inpainter removes the bird and fills in with some realistic background content (d).
	}
	\vskip -0.2in
	\label{fig:removal_methods}
\end{figure}
To compare these two objectives, we randomly chose 50 validation-set images from 52 ImageNet \class{bird} classes and computed their segmentation masks via a pre-trained DeepLab model \cite{chen2017deeplab}.
We found that using the DeepFill-v1 to inpaint the foreground region (\ie our ``deletion'' task) yields realistic samples where the object is removed.
In contrast, using the inpainter to fill in the missing background area \cite{chang2019explaining} yields unrealistic images whose backgrounds contain features (\eg bird feathers or beaks) unnaturally pasted from the object (Fig.~\ref{fig:FIDO_cropped}).
This result motivated us to integrate DeepFill-v1 into MP2 but with the ``deletion'' objective.

\subsection{Inpainter is effective in removing discriminative features}
\label{sec:discriminative}
While removing objects from an image via DeepFill-v1, qualitatively, yields realistic samples, here, we quantitatively test how effective this strategy is in removing target-class discriminative features in comparison with three existing filling methods: (1) zero pixels; (2) random noise; or (3) blurred versions of the original content. Using the same procedure as described in Sec.~\ref{sec:fido}, we randomly sampled 1000 bird images and segmented out the bird in each image.
We filled in the object mask in each image via all four methods (Fig.~\ref{fig:removal_methods}) and compared the results (Table~\ref{table:absence}).
Surprisingly, the blurred and grayed-out images are still correctly classified at 26.4\% and 13.3\% (Table~\ref{table:absence}), respectively, by a pre-trained Inception-v3 classifier \cite{szegedy2016rethinking}, \ie, these perturbed images still contain discriminative features relevant to the target class.
In contrast, only 8.9\% of the inpainted images were correctly classified suggesting that the inpainter removes the discriminative features more effectively.
After the main subject (here, birds) are removed from an image, one would expect the modified image to be perceptually different from the original image (where the bird exists).

\begin{table}[h]
    \vskip -0.2in
	\caption{
		Evaluation of four different filling methods on 1000 random \class{bird} images.
		The Inception-v3 accuracy scores suggest that inpainting the object mask (d) removes substantially more discriminative features relevant to the removed object compared to blurring (b) or graying out (c).
		Perceptually, the inpainted images are also more dissimilar to the corresponding real images according two similarity metrics: MS-SSIM (lower is better) and LPIPS (higher is better).
	}
	\label{table:absence}
	\def\arraystretch{1.2}%
	\begin{center}
		\begin{tabular}{c|c|c|c|c|c}
			\hline
			\multirow{2}{*}{Metrics} & \multicolumn{5}{c}{Filling methods} \\ \cline{2-6} 
			& {\makecell{(a) Real}} & {\makecell{(b) Blur}} & {\makecell{(c) Gray}} & {\makecell{(d) Inpaint}} & {\makecell{(e) Noise}} \\
			\hline
			Inception Acc.(\%)  & {92.30} & {26.40} & {13.30} & {8.90} & {4.40} \\ 
			MS-SSIM & {1.000} & {0.941} & {0.731} & {0.707} & {0.692} \\
			LPIPS & {0.000} & {2.423} &  {3.186} &  {3.208} &  {3.639} \\ 
			\hline
		\end{tabular}
	\end{center}
	\vskip -0.3in
\end{table}

\begin{figure}[h]
	\centering
	\vskip -0.1in
	\includegraphics[width=\columnwidth]{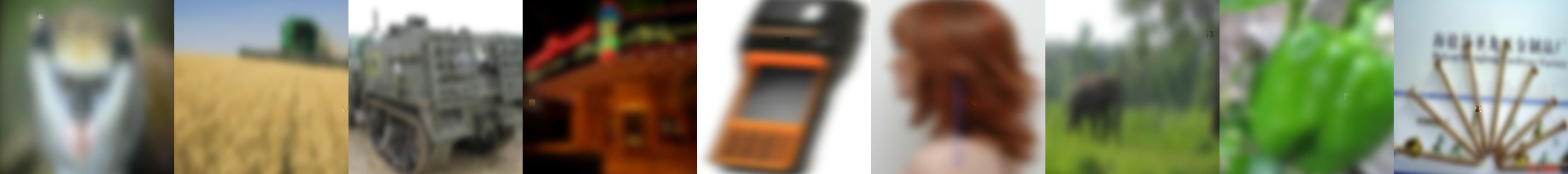}
	{	
		\begin{flushleft}
		    \vskip -0.1in
			\hskip 0.12in 0.047 
			\hskip 0.22in 0.046
			\hskip 0.22in 0.871
			\hskip 0.22in 0.022
			\hskip 0.22in 0.025
			\hskip 0.22in 0.017
			\hskip 0.22in 0.174
			\hskip 0.22in 0.929
			\hskip 0.22in 0.015
		\end{flushleft}
	}
    \vskip -0.1in
	\caption{
	    The target class probability of images do not drop to random guess, \ie $0.001$ for ImageNet, even after perturbing the entire image with a Gaussian blur radius of $\sigma=$10.
	}
	\vskip -0.4in
	\label{fig:mp2_perturbed}
\end{figure}

Here, we evaluate how each of the four in-filled images $\bar{\vx}$ (where the bird has been removed) is perceptually dissimilar to the original image $\vx$ by measuring the MS-SSIM and LPIPS \cite{zhang2018unreasonable} scores between every pair ($\vx$, $\bar{\vx}$).
Across both metrics, the inpainted images are consistently more dissimilar from the real images compared to the blurred and grayed-out images.
Note that in all three quantitative metrics, the inpainted images are the closest to the noise-filled images (Table~\ref{table:absence}d--e) despite being substantially more realistic (Fig.~\ref{fig:removal_methods}).
Furthermore, the problem with using blurring as a perturbation operation is explicitly seen in cases where attribution maps covers the entire image.
This is because for some inputs even blurring out the entire image does not result in a significant probability (Fig.~\ref{fig:mp2_perturbed}).
Across the set of 2000 images, the average confidence score on blurring the entire image was $0.3198$.

\subsection{Are explanations by G-methods more accurate?}
\label{sec:faithful}
While there are currently no established ground-truth datasets to evaluate the correctness of an attribution map, prior research often assessed correctness via three common proxy metrics: (1) the object localization task \cite{zhou2016learning}; (2) the deletion task \cite{Petsiuk2018rise}; (3) the saliency metric \cite{dabkowski2017real}.
Here, we ran 8 algorithms on the ImageNet and Places365 datasets using the default hyperparameters (Sec.~\ref{sec:methods}).
The heatmaps are then upsampled to the full image resolution for evaluation on all three measures above.

\begin{table*}[h]
    \vskip -0.25in
	\caption{
		Localization errors (lower is better) for all attribution methods on ImageNet. 
		Naively taking the whole image as a bounding box yields an error of 38.56\% (baseline).
		MP2-G outperformed all methods including MP, MP2 and a related FIDO-CA \cite{chang2019explaining}.
	}
	\label{tab:IOU_metric}
	\def\arraystretch{1.22}%
	\begin{center}
	\vskip -0.05in
	\begin{tabular}{c|c|c|c|c|c|c|c|c} 
		\hline
		Baseline & SP \cite{zeiler2014visualizing} & SP-G & LIME \cite{ribeiro2016should} & LIME-G & MP \cite{fong2017interpretable} & MP2 & MP2-G & FIDO-CA \cite{chang2019explaining} \\
		\hline
		39.7\% & 41.9\% & \textbf{38.95\%} & 28.05\% & \textbf{26.55\%} & 29.35\% & 24.4\% & \textbf{24.03\%} & 27.9\% \\
		\hline
	\end{tabular}
	\end{center}
	\vskip -0.2in
\end{table*}

\begin{figure}[ht]
	\centering
	{
		\begin{flushleft}
			\hskip 0.25in (a) Real + BB
			\hskip 0.05in (b) MP2-G
			\hskip 0.14in (c) SP \cite{zeiler2014visualizing}
			\hskip 0.12in (d) LIME \cite{ribeiro2016should}
			\hskip 0.08in (e) MP \cite{fong2017interpretable}
			\hskip 0.17in (f) MP2
		\end{flushleft}
	}
	\vskip -0.14in
	\includegraphics[width=0.90\linewidth]{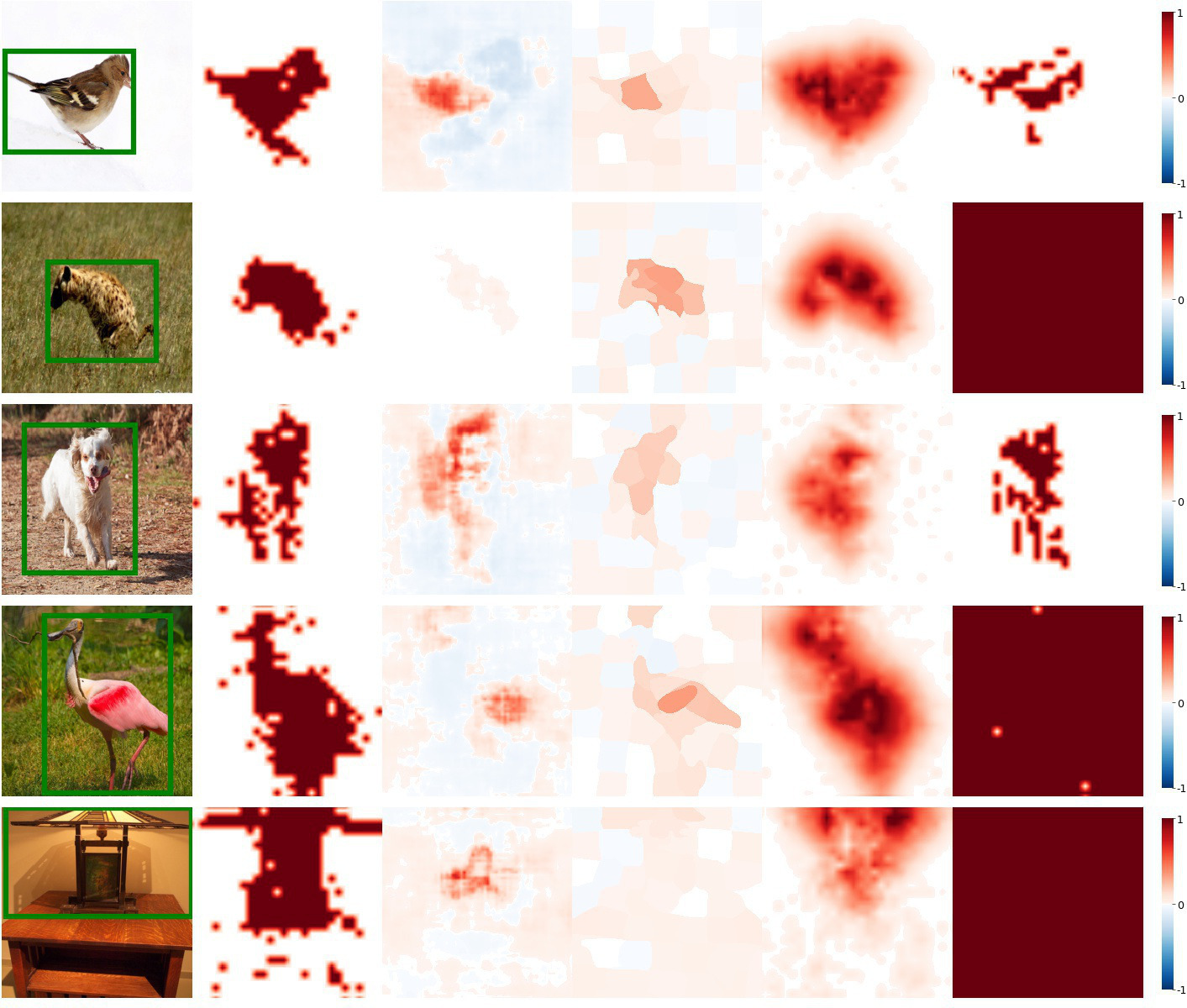}
	\caption{
	    MP2-G results in attribution maps that localize the objects accurately compared to other perturbation-based methods.
	    From left to right, in each row, we show a real ImageNet image with its ground-truth bounding box (BB) (a), attribution maps from the proposed MP2-G (b) and other existing methods (c--f). 
	    Images are randomly chosen. 
	    For qualitative evaluation, Figs.~\ref{fig:IOU_LIME}-\ref{fig:IOU_occlusion} show a set of heatmaps and their derived BB's.
	}
	\label{fig:wow_figure}
	\vskip -0.25in
\end{figure}
\noindent \textbf{Object localization}
Zhou et al. \cite{zhou2016learning} proposed to evaluate heatmaps by localizing objects in the ImageNet images, which often contain a single object of a known class.
We followed the localization procedure in \cite{fong2017interpretable} for the ImageNet dataset. 
That is, for each algorithm, we derived multiple bounding boxes per heatmap by thresholding it at different values of $t=\alpha \mu_{max}$, where $\mu_{max}$ is the maximum intensity in the heatmap and $\alpha\in[0:0.05:0.95]$. 
For each $\alpha$, we computed the Intersection over Union (IoU) score between a derived bounding box and the ImageNet ground-truth.
The object localization error was calculated by thresholding each IoU score at $0.5$ and averaging them across the number of images.
For each method, we chose the best $\alpha^*$ that yielded the lowest error on a held-out set of 1000 ImageNet images (Table~\ref{tab:IOU_metric}).
\textbf{
We found that our generative version of the attribution algorithms outperformed their respective counterparts and MP2-G outperformed FIDO-CA\footnote{We produced FIDO-CA results using the code provided by the authors \cite{chang2019explaining}. See Sec.~\ref{sec:codeFIDO} for more details.} (Table~\ref{tab:IOU_metric}).
}
Among the 8 methods, MP2-G obtained the lowest error of $24.03\%$.
Qualitatively, MP2-G generates attribution maps that are more localized to the objects in the image (Fig.~\ref{fig:wow_figure}).\\

\begin{table*}[h]
    \vskip -0.35in
	\caption{
		\textbf{Deletion metric} (lower is better): SP-G, LIME-G, and MP2-G outperformed their counterparts on both ImageNet and Places365 datasets.
		G-methods also outperformed a baseline (here, random attribution maps).
	}
	\label{tab:DELINS_metric}
	\def\arraystretch{1.2}%
	\begin{center}
	\vskip -0.05in
	\begin{tabular}{c|c|c|c|c|c|c|c|c|c}
		\hline
		Dataset     & Baseline & SP\cite{zeiler2014visualizing}    & SP-G            & LIME\cite{ribeiro2016should}   & LIME-G          & MP\cite{fong2017interpretable} & MP2 & MP2-G & FIDO-CA \cite{chang2019explaining}  \\ \hline
		ImageNet  & 0.2083 & 0.1996 & \textbf{0.1769} & 0.1355 & \textbf{0.1171} & 0.1654 & 0.1530 & \textbf{0.1311} & 0.1638\\ \hline
		Places365 & 0.2151 & 0.2560 & \textbf{0.1944} & 0.1919 & \textbf{0.1582} & 0.2014 & 0.1980 & \textbf{0.1871} & 0.1987 \\ \hline
	\end{tabular}
	\end{center}
	\vskip -0.25in
\end{table*}

\noindent \textbf{Deletion metric~} 
Intuitively, if the attributions in an explanation correctly reflect the importance of input pixels, removing the input pixels of highest attributions should cause a substantial probability drop.
The deletion metric \cite{Petsiuk2018rise} measures the area under the curve of the target-class probability as we gradually zero out input pixels of the highest attributions in descending order.
The deletion scores are widely used to compare attribution methods \cite{arras2017relevant,wagner2019interpretable,hooker2019benchmark,samek2016evaluating} \ie, lower deletion scores are considered more accurate.
Here, we evaluated all 8 methods via the code released by \cite{Petsiuk2018rise} where the authors knocked out $224\times8$ pixels at a time.
Similar to object localization results, we observed a consistent trend: \textbf{Across both ImageNet and Places365, our G-methods outperformed their counterparts while MP2-G outperformed all algorithms} (Table~\ref{tab:DELINS_metric}).\\

\noindent \textbf{Saliency metric}
Dabkowski et al. \cite{dabkowski2017real} proposed that if an explanation is accurate then the most salient patch in an image (derived from the attribution map) should have a high prediction score.
That is, we took the smallest rectangular patch derived from thresholding the attribution map using an $\alpha^{*}$ which yielded the least salient metric score on a held-out dataset of 1000 images (similar to the object localization task).
The saliency metric is then defined as $\log\big({\max({a, 0.05})}\big) - \log({s(\vx_p)})$
where $a$ is the ratio of the patch size over the image size and $s(\vx_p)$ is the classification probability for the patch $\vx_p$ upsampled to the full image size.
A lower saliency score indicates a more accurate explanation.
\textbf{
On both ImageNet and Places365, SP-G and MP2-G obtained lower scores than their counterparts while LIME-G was on par with LIME (Table~\ref{tab:SM_metric}).
	MP2-G outperformed all its baselines, \ie MP, MP2, and FIDO-CA.
}
We hypothesize that the difference between LIME vs. LIME-G is small because they operate at the superpixel level and most salient \emph{pixels} might fall in common \emph{superpixels} across their respective explanations.
Refer to Fig.~\ref{fig:alpha_performance} for the localization error and saliency metric scores for different $\alpha$'s on the held-out set of 1000 images.

\begin{table*}[h]
    \vskip -0.25in
	\caption{
		\textbf{Saliency metric} (lower is better):
		On both ImageNet and Places365, while LIME and LIME-G performed on-par, SP-G and MP2-G consistently outperformed their counterparts (MP2-G outperformed its baselines: FIDO-CA, MP and MP2).
		The baseline was calculated using a random attribution map.
	}
	\label{tab:SM_metric}
	\def\arraystretch{1.2}%
	\begin{center}
	\vskip -0.05in
	\begin{tabular}{c|c|c|c|c|c|c|c|c|c} 
		\hline
		Dataset & Baseline & {SP \cite{zeiler2014visualizing}} & {SP-G} & {LIME\cite{ribeiro2016should}} & {LIME-G} & {MP\cite{fong2017interpretable}} & {MP2} & {MP2-G} & {FIDO-CA \cite{chang2019explaining}} \\
		\hline
		ImageNet & 0.3294 & {0.3774}	& \textbf{0.3122} & {0.1191} &	{0.1159} & {0.1182} & {0.0890} & \textbf{0.0540} & {0.1071} \\
		\hline
		Places365 & 1.117 & 1.1311 & \textbf{1.1148} & 0.9597 & 0.9568 & 1.0413 & 0.9331 & \textbf{0.9156} & 0.9263 \\
		\hline
	\end{tabular}
	\end{center}
	\vskip -0.35in
\end{table*}

\subsection{Are G-methods more robust to hyperparameter changes?}
\label{sec:sensitivity}

Machine learning methods are highly sensitive to hyperparameters, contributing to the reproducibility crisis \cite{hutson2018crisis}.
Similarly, perturbation-based attribution methods were recently found to be highly sensitive to common hyperparameters \cite{bansal2020sam}.
Such sensitivity poses a huge challenge in (1) evaluating the explanations; and (2) building trust with end users \cite{doshi2017towards}.
Our hypothesis is that heuristically-perturbed samples are often far from the true data distribution and thus contribute to the hyperparameter sensitivity of heatmaps.
Here, we test whether our generative methods are more robust to hyperparameter changes than their original counterparts.

\subsec{Similarity metrics and Image sets~}
Following \cite{adebayo2018sanity,bansal2020sam}, we used three metrics from scikit-image \cite{scikit-image} to measure the similarity of heatmaps:
the Structural Similarity Index (SSIM), the Pearson correlation of the histograms of oriented gradients (HOGs), and the Spearman rank correlation. 
We upsampled all heatmaps to the full image size before feeding them into the similarity metrics.
We performed the test on a set of 1000 random images from both ImageNet and Places365.	
\begin{figure*}[t]
	\begin{tabular}{cc}
		\includegraphics[width=0.33\linewidth]{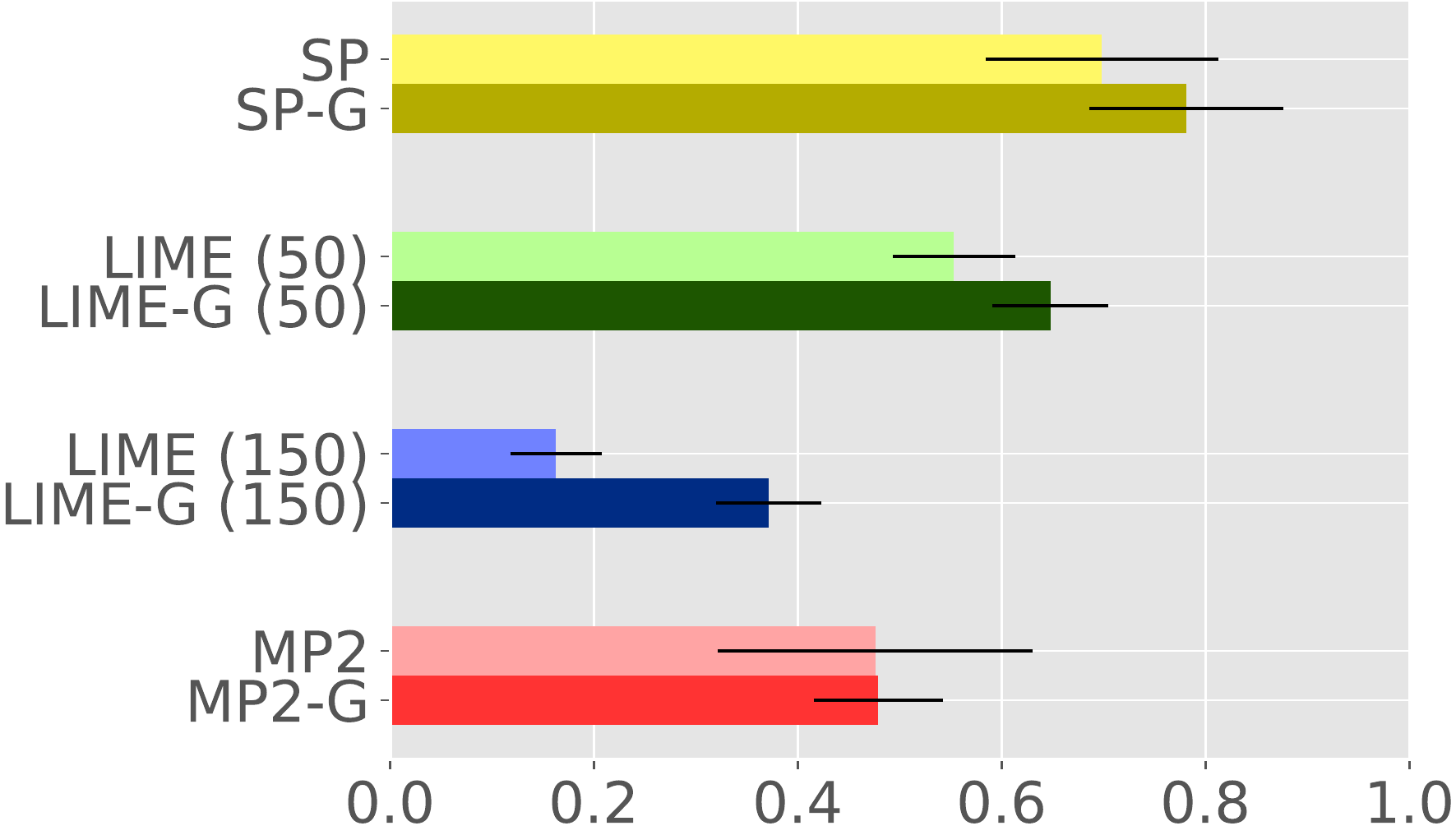}  
		\includegraphics[width=0.33\linewidth]{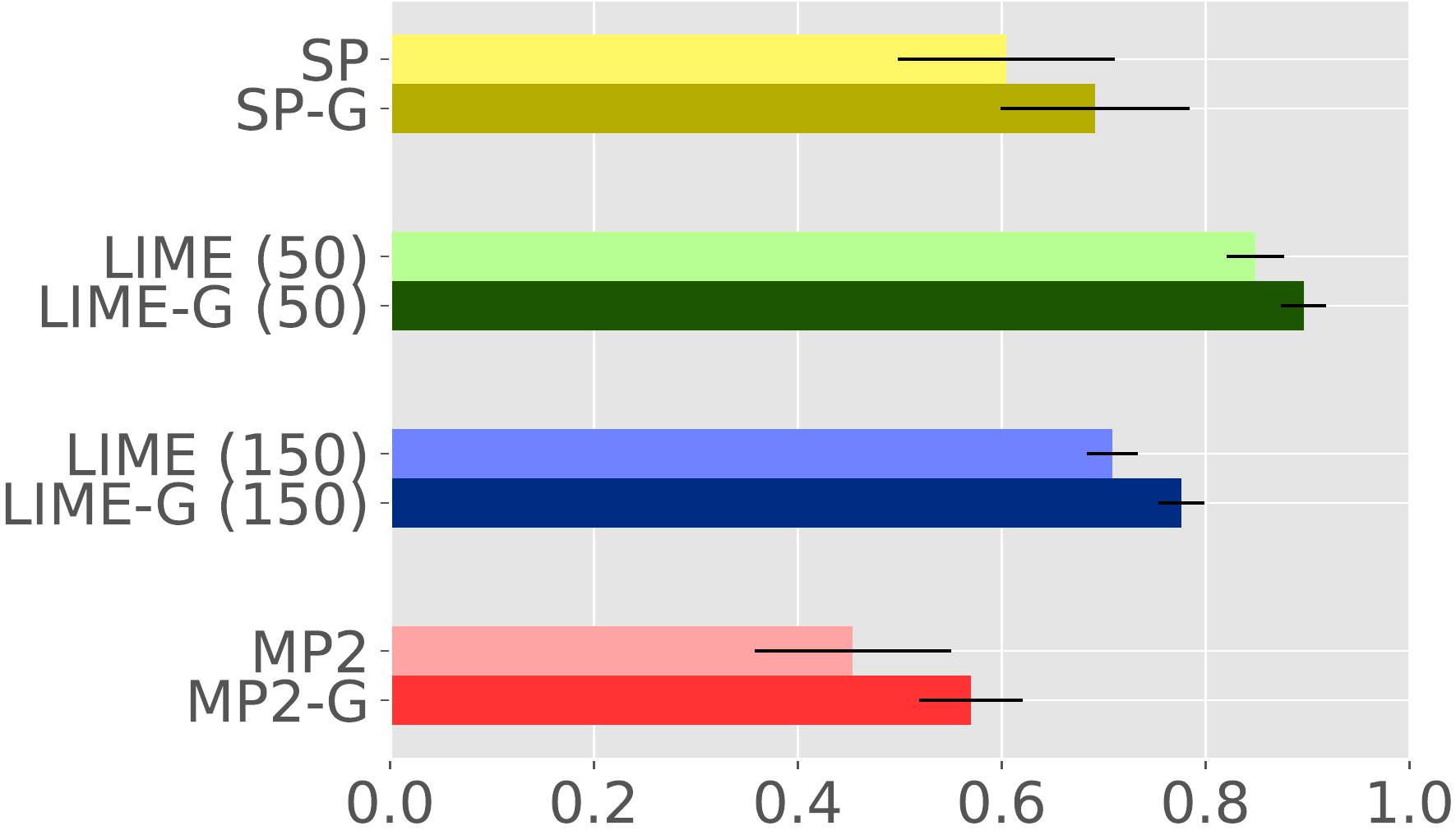}   
		\includegraphics[width=0.33\linewidth]{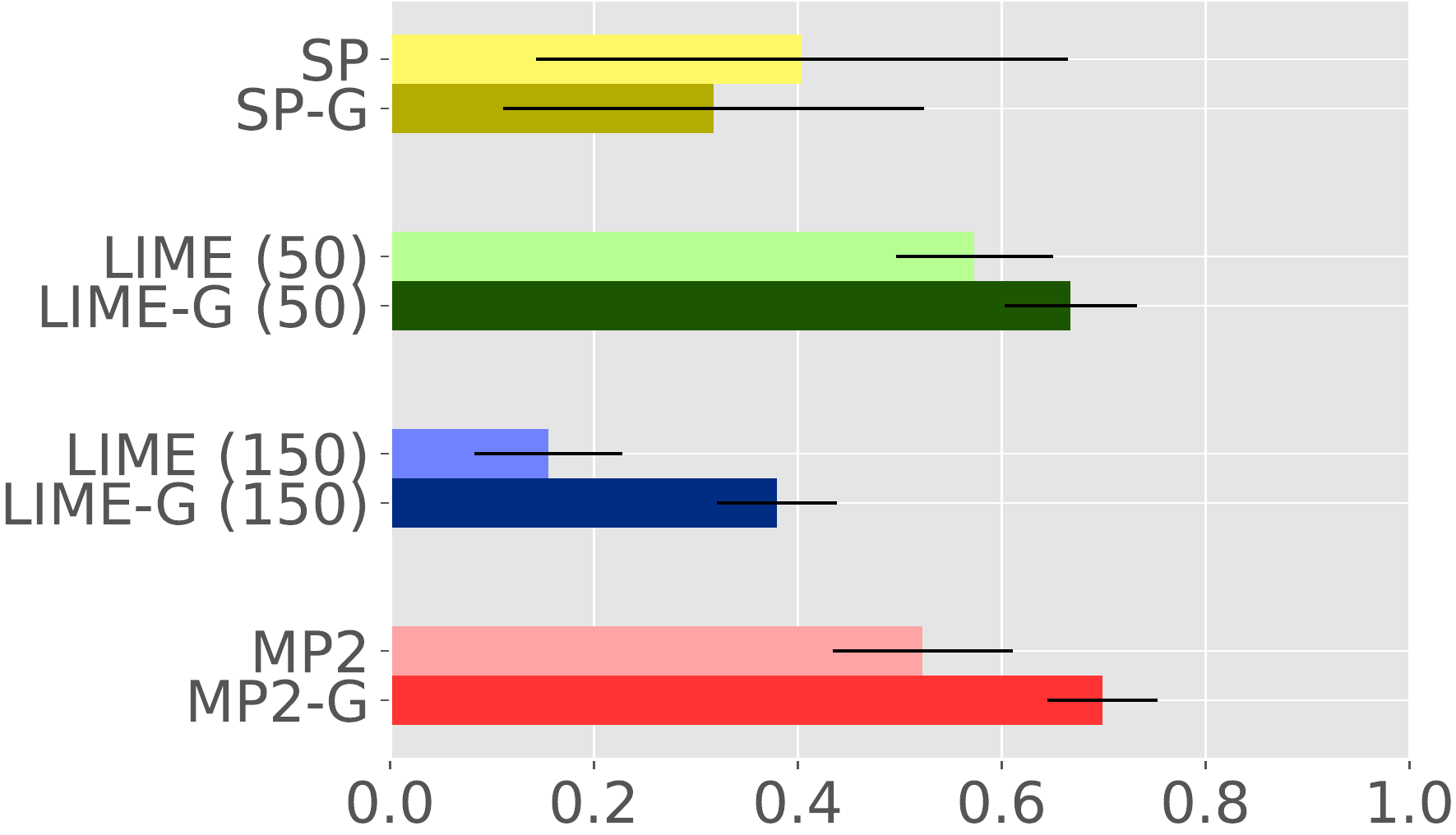}\\
	\end{tabular}
	{	
		\begin{flushleft}
			\hskip 0.6in (a) SSIM
		    \hskip 0.2in (b) Pearson correlation of HOG features
	        \hskip 0.02in (c) Spearman rank correlation
		\end{flushleft}
	}
	\vskip -0.1in
	\caption{
		Error plots for SSIM (a), Pearson correlation of HOG features (b), and Spearman rank correlation (c) scores obtained from 1000 random ImageNet images (higher is better). 
		G-methods are more robust than their counterparts (dark vs light bars).
		LIME-G, in particular, is robust than LIME on both low and high resolutions \ie $S \in \{50, 150\}$ (green and blue bars).
		The same trends were also observed on the Places365 dataset (Fig.~\ref{fig:sensitivity_score_Places365}). 
		The exact numbers are reported in Table~\ref{tab:sensitivity_ImageNet}.
	}
	\label{fig:sensitivity_score_ImageNet}
	\vskip -0.15in
\end{figure*}

\subsec{SP sensitivity across patch sizes~}
It remains a question how to choose the patch size in the SP algorithm because changing it can change the explanation radically \cite{zintgraf2017visualizing}.
Therefore, we compare the sensitivity of SP and SP-G when sweeping across 5 patch sizes $p \times p$ with stride $3$ where $p \in \{5, 17, 29, 41, 53\}$.
We chose this set to cover the common sizes that have been used in the literature.
For each input image, we obtained $k=5$ heatmaps (\ie each corresponds to a patch size) and then measured the similarity among all $k(k-1)/2 = 10$ possible pairs.

\subsec{LIME sensitivity across random batches of samples~}
LIME randomly samples $N$ perturbed images $\{\bar{\vx}^i\}_N$ and uses them to fit a heatmap.
Therefore, we compared the sensitivity of LIME and LIME-G across 5 random batches of $N = 500$ perturbation samples.
That is, for each input image among the 1000, we generated $k=5$ heatmaps and computed the similarity among all 10 possible pairs.
We ran this experiment for a small and a large heatmap resolution \ie two numbers of superpixels $S \in \{50, 150\}$ while keeping all other hyperparameters constant.

\subsec{MP2 sensitivity across mask sizes~}
Because optimizing a mask at a high resolution is prohibitively slow, Fong et al. \cite{fong2017interpretable} used an MP mask of size $28\times28$ and upsampled it to the image size when applying the blur operator on the input image.
Therefore, the mask size is a hyperparameter of MP2 and MP2-G.
Here, we compare the sensitivity by sweeping across the three mask sizes where $D \in \{28, 56, 112\}$.
We re-ran each algorithm three times on each input image to yield three heatmaps and computed the average pairwise similarity scores from all possible pairs of heatmaps.
\subsubsection{Results} 
First, we found that all 6 algorithms produce inconsistent explanations across the controlled hyperparameters (Fig.~\ref{fig:sensitivity_score_ImageNet}; all scores are below 1).
That is, LIME heatmaps can change as one simply changes the random seed!
However, LIME-G is consistently more robust than LIME across all metrics and superpixel settings (Figs.~\ref{fig:sensitivity_score_ImageNet}~\&~\ref{fig:LIME_superpixel_sens}).
Across the patch sizes, SP-G is also consistently more robust than SP (Fig.~\ref{fig:sensitivity_score_ImageNet}a--b; light vs. dark yellow).
SP-G and SP performed on par with high standard deviations under the Spearman rank correlation (Fig.~\ref{fig:sensitivity_score_ImageNet}c).
Across the optimization mask size, MP2-G is consistently more robust than MP2 (Fig.~\ref{fig:sensitivity_score_ImageNet}; light vs. dark red).

\section{The inner-workings of generative attribution methods}
Here, we further explain why our G-methods are both more (1) accurate in localizing objects (Sec.~\ref{sec:faithful}) and (2) robust to hyperparameter changes (Sec.~\ref{sec:sensitivity}).

\begin{figure}[ht]
	{	
		\begin{flushleft}
		    \vskip -0.1in
			\hskip 0.25in $9$
			\hskip 0.25in $24$
			\hskip 0.20in $36$
			\hskip 0.20in $44$
			\hskip 0.20in $53$
		\end{flushleft}
	}
	\vskip -0.2in
	\subcaptionbox{Perturbation samples \& heatmaps (rightmost)\label{fig:SP_samples}}%
	[.49\textwidth]{\includegraphics[width=0.51\columnwidth]{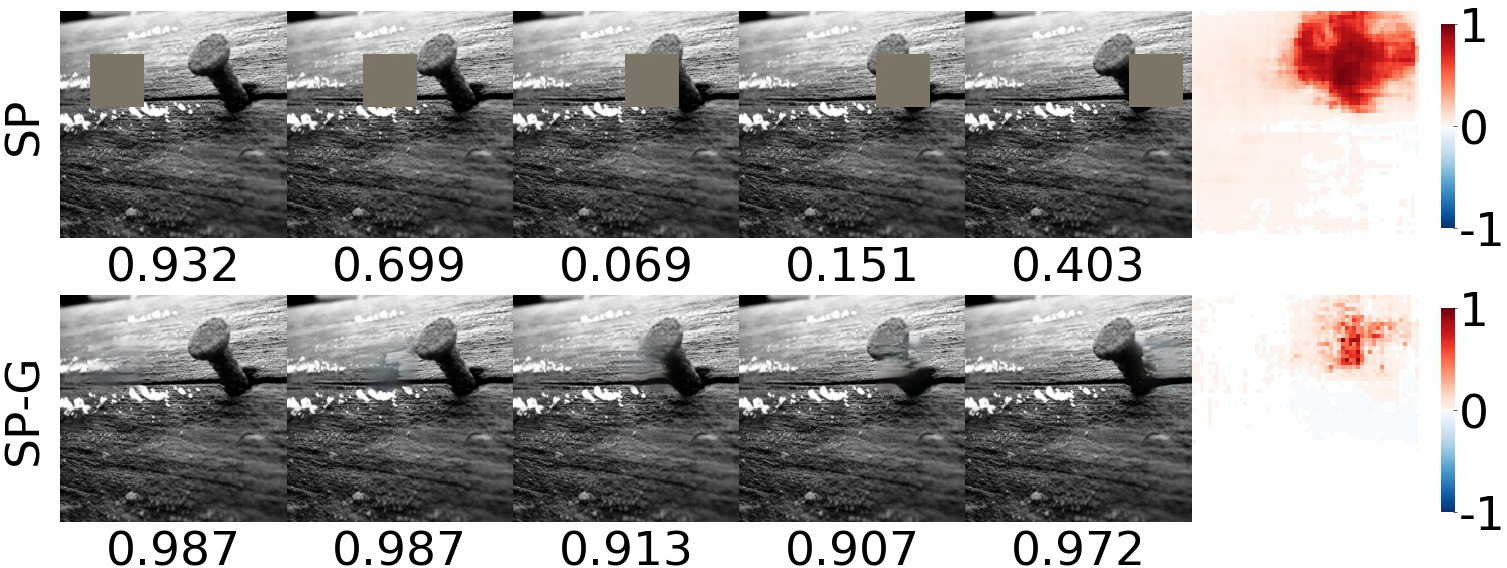}}
	\subcaptionbox{GT-class probability over patch locations\label{fig:SP_lineplot}}%
	[.49\textwidth]{\includegraphics[width=0.45\columnwidth]{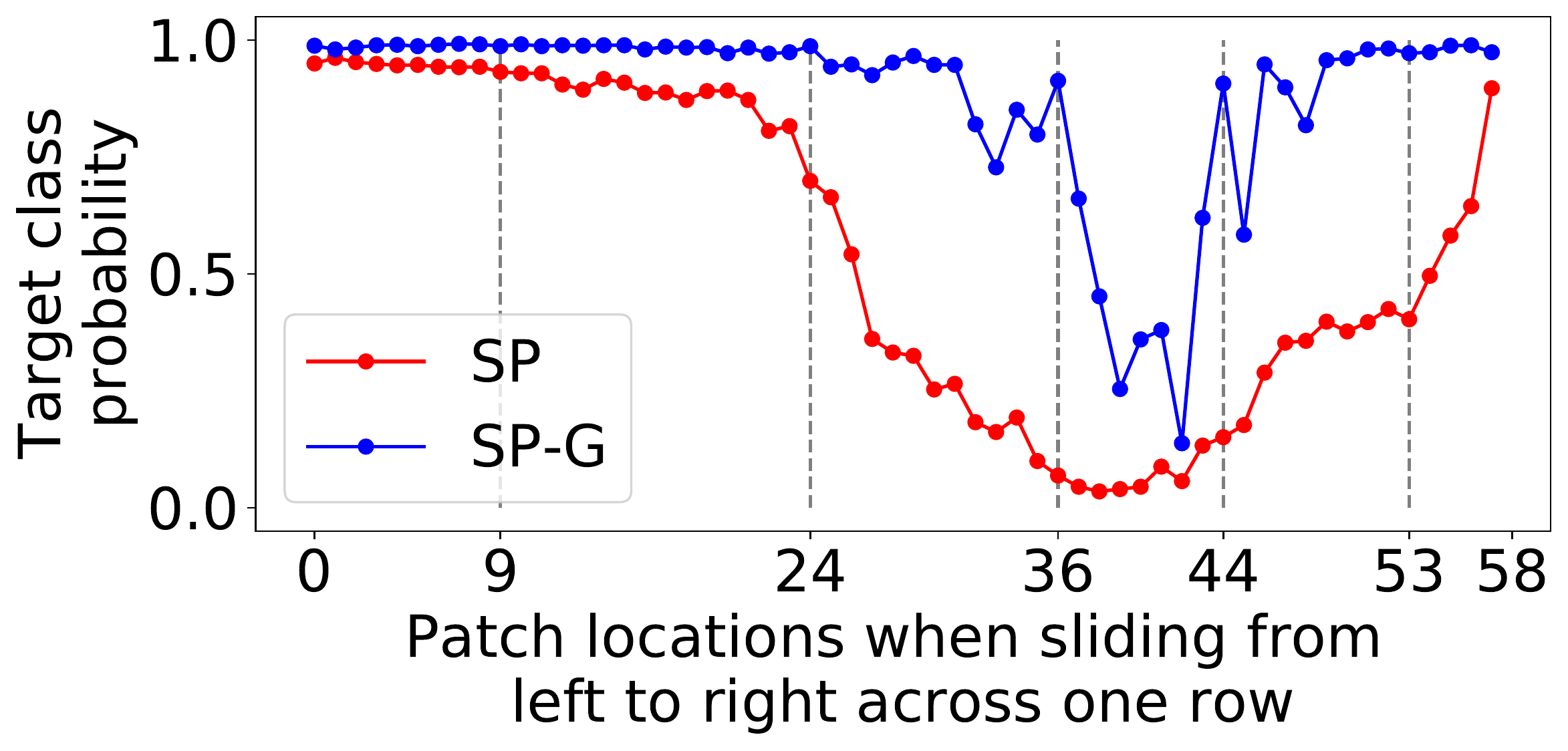}}
	\caption{
		We ran SP and SP-G using a $53\times53$ patch on a \class{nail} class image.
		Here are the perturbation samples from both methods when the patch is slided horizontally across a row at 5 locations $\{9, 24, 36, 44, 53\}$ (a); and their respective target-class probabilities (b).
		SP-G samples are more realistic than SP and its heatmap localizes the object accurately (a). 
		That is, the probabilities for SP-G samples are more stable and only substantially drop when the patch covers the object (blue vs. red).
		See Fig.~\ref{fig:SP_teaser_perturbed} for more examples.
	}
	\label{fig:SP_teaser}
	\vskip -0.35in
\end{figure}

\begin{figure}[ht]
	\centering
    \includegraphics[width=\linewidth]{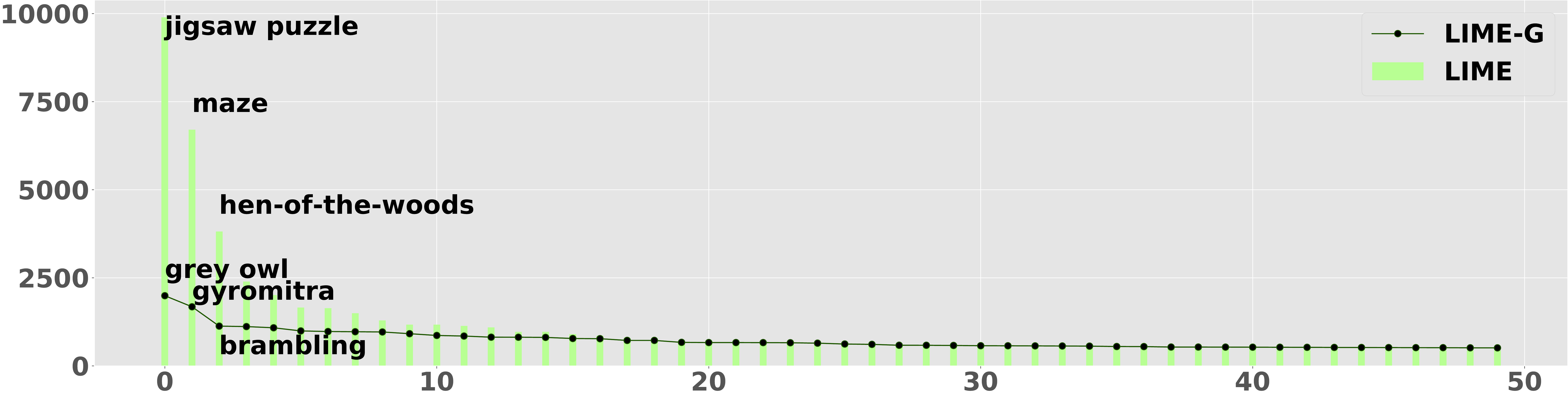}
	\caption{
		We ran LIME and LIME-G on 200 images, each run has 500 intermediate perturbation samples.
		Here, for LIME (light green) and LIME-G (dark green) samples, we show a histogram of the top-1 predicted class labels for all $200$ runs $\times 500$ samples = 100,000 images.
		LIME perturbed samples are highly biased towards few \class{jigsaw~puzzle}, \class{maze} classes (left panel), which is somewhat intuitive given the gray-masked images (see Figs.~\ref{fig:LIME_ImageNet_top_sensitivity_1}--\ref{fig:LIME_ImageNet_bottom_sensitivity_2}).
		In contrast, the histogram of LIME-G samples are almost uniform.
		\textbf{x-axis:} For visualization purposes, we sorted the top-1 labels and showed only first 50 labels.
		See Fig.~\ref{fig:LIME_histogram_full} for an expanded version of the figure. 
	}
	\label{fig:LIME_histogram}
	\vskip -0.2in
\end{figure}
\subsection{More accurate object localization: A case study of SP-G}
We found that as the gray patch of SP is slided from left to right across the input image (Fig.~\ref{fig:SP_samples}; top), the target-class probability gradually decreases and approaches 0 when the patch occludes most of the object (Fig.~\ref{fig:SP_lineplot}; red line).
Notably, the probability even drops when the patch is far outside the object region (Fig.~\ref{fig:SP_lineplot}; red line within $[0, 24]$) due to SP's unrealistic grayish samples.
Hence, the probability distributions by SP often yield a large blob of high attributions around the object in the heatmap (Fig.~\ref{fig:SP_samples}; top-right).
In contrast, the inpainted samples of SP-G often keep the probability variance low except when the patch overlaps with the object (Fig.~\ref{fig:SP_lineplot}; blue vs. red), yielding heatmaps that are more localized towards the object (Fig.~\ref{fig:SP_samples}; bottom).
Across 1000 random ImageNet images, we found that the average probability change when the SP $53\times53$ patch is outside the object bounding box is $\sim$2.1$\times$ higher than that of SP-G (\ie 0.09 vs. 0.04).
In sum, our observations here are consistent with the findings that G-methods obtained lower localization errors than the original counterparts.
\subsection{More robust heatmaps: A case study of LIME-G}
\label{sec:explain_LIME}
\begin{figure}[ht]
	\subcaptionbox{5 LIME perturbation samples\label{fig:LIME_samples}}%
	[0.49\linewidth]{\includegraphics[width=0.49\columnwidth]{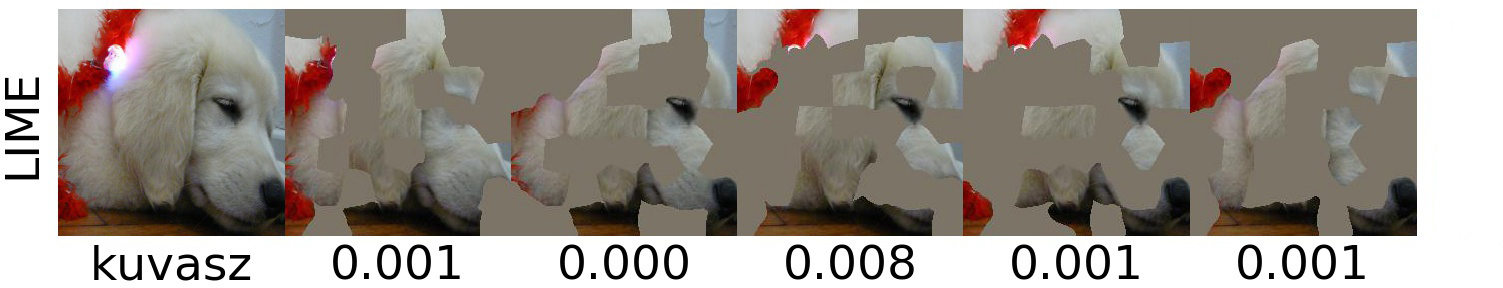}}
	\subcaptionbox{5 LIME heatmaps using five random seeds\label{fig:LIME_heatmaps}}%
	[0.49\linewidth]{\includegraphics[width=0.49\columnwidth]{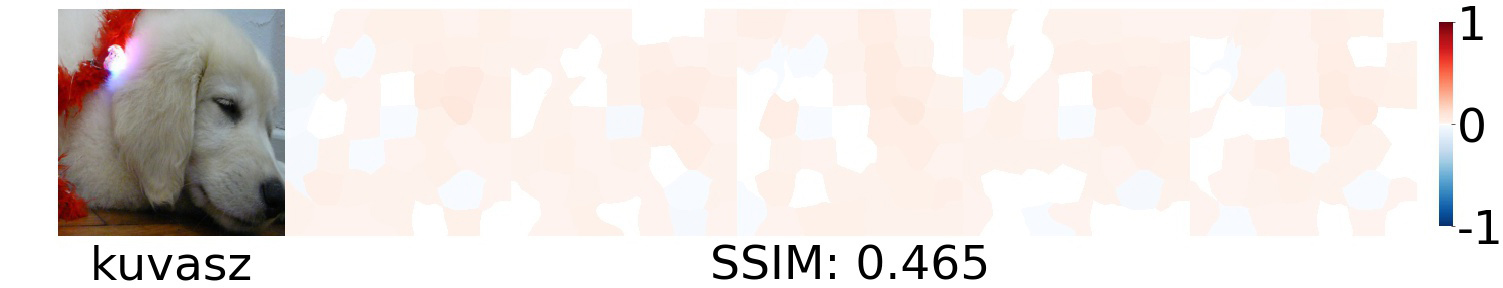}}\\
	\subcaptionbox{5 LIME-G perturbation samples \label{fig:LIME-G_samples}}%
	[0.49\linewidth]{\includegraphics[width=0.49\columnwidth]{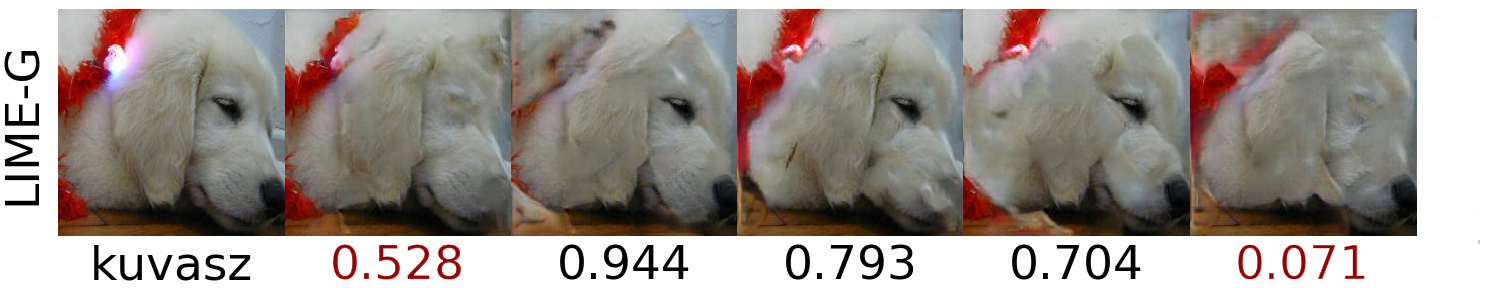}}
	\subcaptionbox{5 LIME-G heatmaps using five random seeds\label{fig:LIME-G_heatmaps}}%
	[0.49\linewidth]{\includegraphics[width=0.49\columnwidth]{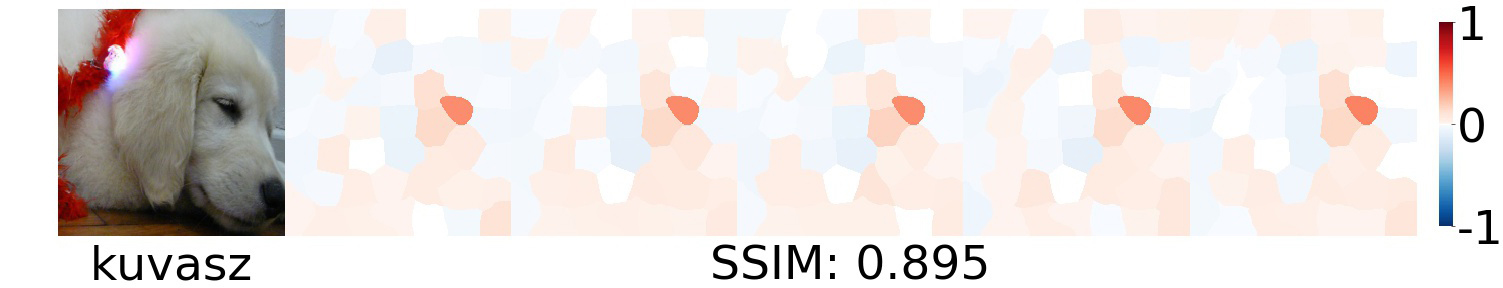}}
	\caption{
		In Sec.~\ref{sec:sensitivity}, we compared the robustness of LIME vs. LIME-G heatmaps when running using 5 different random seeds.
		This is an example where LIME-G heatmaps are more consistent than LIME's (d vs. b).
		While LIME grayish samples (a) are given near-zero probabilities, LIME-G samples (here, inpainted using the same masks as those in the top row) are often given high probabilities except when the \class{kuvasz} dog's eye is removed (c).
		LIME-G consistently assign attributions to the dog's eye (d) while LIME heatmaps appear random (b).
		The top-1 predicted labels for 4 out of 5 LIME samples (a) here are \class{paper~towel}.
	}
	\label{fig:LIME_teaser}
	\vskip -0.2in
\end{figure}

Here, we provide insights for why LIME-G produced heatmaps that are more consistent than LIME across 5 random batches of samples.
We observed that the top-1 predicted labels of $\sim$20.5\% of the LIME grayish perturbation samples (\eg Fig.~\ref{fig:LIME_samples}) were from only three classes \{~\class{jigsaw~puzzle}, \class{maze}, \class{hen}-\class{of}-\class{the}-\class{wood}~\} whereas the same top-1 label distribution for LIME-G samples was almost uniform (see Fig.~\ref{fig:LIME_histogram} for more details).
Due to their similar grayish, puzzle-like patterns, many LIME samples across images from different classes (e.g. \class{dogs} or \class{nail}) are still classified into the same label!
Relatedly, we observed that a LIME perturbation sample is often given a near-zero probability score \emph{regardless of what input feature is being masked out} (Fig.~\ref{fig:LIME_samples}).
Therefore, when fitted to $N$ samples, where $N$ is often too small w.r.t. the total $2^S$ possible samples, the heatmap appears random and changes upon a new set of random masks (Fig.~\ref{fig:LIME_heatmaps}).

In contrast, for LIME-G samples, the probabilities consistently drop when some discriminative features (\eg the \class{kuvasz} dog's eye in Fig.~\ref{fig:LIME-G_samples}) are removed. 
This phenomenon yields heatmaps that are more consistently localized around the same input features across different random seeds (Fig.~\ref{fig:LIME-G_heatmaps}).
Our explanation also aligns with the finding that when the number of superpixels $S$ increases from $50$ to $150$ (while the sample size remains at $N=500$), the sensitivity gap between LIME vs. LIME-G increases by $\sim$3 times (Fig.~\ref{fig:sensitivity_score_ImageNet}a; gap between green bars vs. gap betwen blue bars).
See Figs.~\ref{fig:LIME_ImageNet_top_sensitivity_1}--\ref{fig:LIME_Places365_bottom_sensitivity_2} for qualitative examples of when LIME-G is more robust than LIME and vice-versa.
Quantitatively, we found that the image distribution where LIME-G showed superior robustness over LIME \emph{across all three similarity metrics} mostly contains images of scenes, close-up or tiny objects.
In contrast, LIME is more robust than LIME-G on images of mostly birds and medium-sized objects (See Sec.~\ref{sec:inputs} for more details).

\section{Discussion and Conclusion}
MP2-G outperforming FIDO-CA consistently on all accuracy metrics confirms that the ``deletion'' objective is more appropriate for MP2 when incorporating generative inpainters.
Additionally, discretizing and removing the hyperparameters of the original MP formulation aid in generating attribution maps that achieve better results across localization error, deletion, and saliency metric scores.

Integrating a state-of-the-art inpainter into three representative attribution methods consistently yielded explanations that are (1) more accurate based on three metrics; (2) more robust to hyperparameter changes; and (3) based on more plausible counterfactuals.
Our results suggest that harnessing generative models to generate synthetic interventions (here, removal of input features) is a promising direction for future causal explanation methods.

\subsection*{Acknowledgments}
We thank Qi Li and Michael Alcorn for helpful feedback. Especially, we thank Naman Bansal for valuable discussions, feedback on the final draft, and an important pointer to a closely related work.
AN is supported by the National Science Foundation under Grant No. 1850117, Amazon Research Credits, Auburn University, and donations from Nvidia.



\newpage
\clearpage
\bibliographystyle{splncs}
\bibliography{references}


\appendix
%
%
%

\newpage
\onecolumn

\renewcommand{\thesection}{A\arabic{section}}
\renewcommand{\thesubsection}{\thesection.\arabic{subsection}}

\newcommand{\beginsupplementary}{%
	\setcounter{table}{0}
	\renewcommand{\thetable}{S\arabic{table}}%
	\setcounter{figure}{0}
	\renewcommand{\thefigure}{S\arabic{figure}}%
	\setcounter{section}{0}
}
\newcommand{\suptitl}{Supplementary Information for:\\ \titl}
\newcommand{\suptitlrunning}{Supplementary Information for: \titl}

\beginsupplementary
\title{\suptitl}

\section{Implementation details of FIDO-CA}
\label{sec:codeFIDO}
For all the results of FIDO-CA, we followed the implementation details in the code released on Github \url{https://github.com/zzzace2000/FIDO-saliency} by the authors \cite{chang2019explaining}.
FIDO-CA was ran using the ``preservation'' objective in conjunction with the DeepFill-v1 \cite{yu2018generative} inpainter that we also harnessed in this paper.
For the optimization, we used Adam optimizer with a learning rate of $0.05$ and a regularization coefficient of 0.001.
A coarse $56\times56$ mask was optimized using a ResNet-50 classifier for the ImageNet-S and Places365-S datasets respectively.
The mask was finally upsampled to the full image size, \ie, $224\times224$, using bilinear interpolation.

\section{LIME-G is more robust than LIME on images of scenes, close-up and tiny objects}
\label{sec:inputs}
We have shown that LIME-G is more robust than LIME consistently on all 3 different similarity metrics (see Sec.~\ref{sec:sensitivity} in the main text).
Here, we aim to understand the image distributions where LIME-G was more robust than LIME and vice versa.

\begin{figure}[ht]
	\centering
	\includegraphics[width=0.7\linewidth]{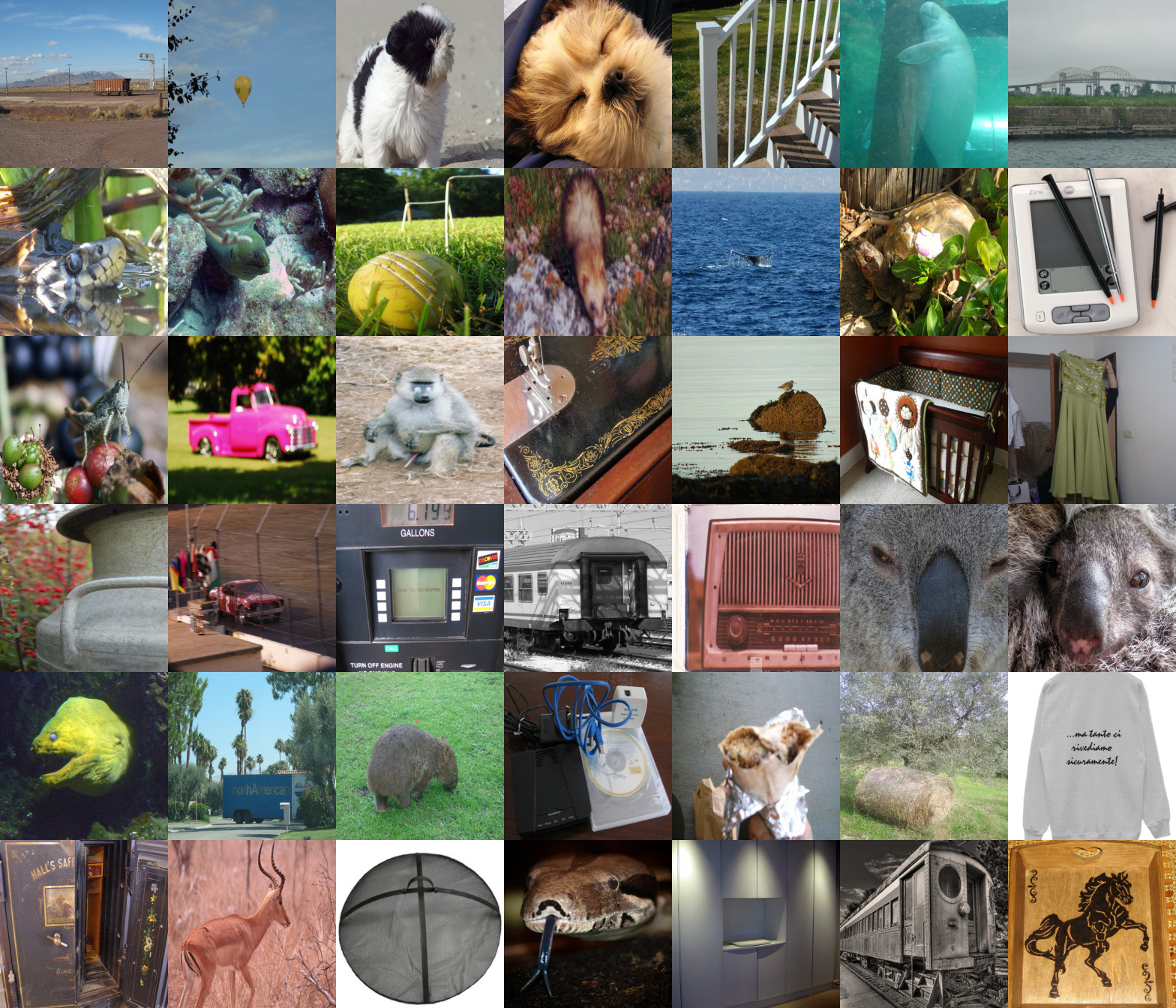}
	{	
		\begin{flushleft}
			\hskip 0.3in Images where LIME-G \underline{outperformed} LIME across all three sensitivity metrics
		\end{flushleft}
	}
	\includegraphics[width=0.7\linewidth]{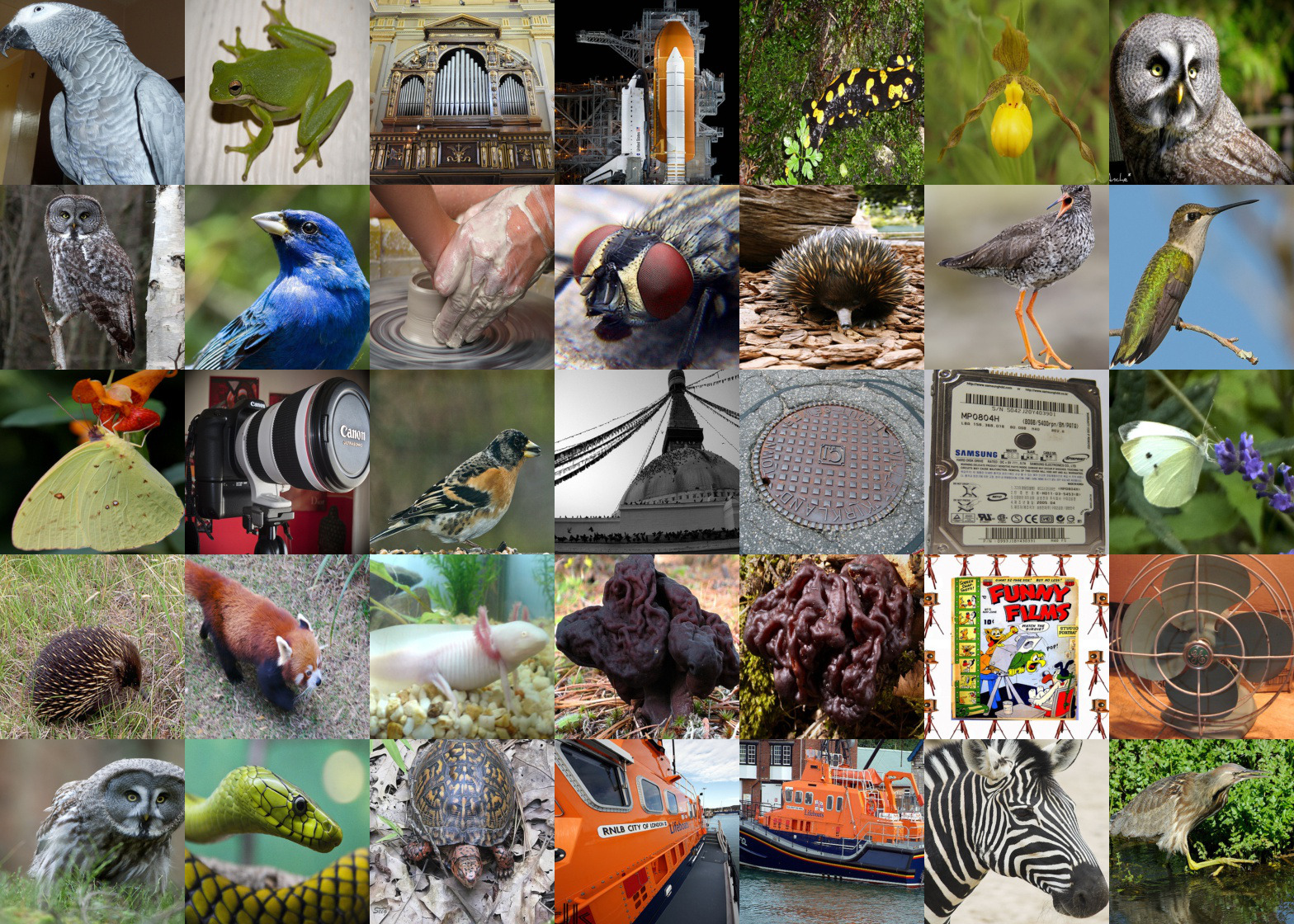}
	{	
		\begin{flushleft}
			\hskip 0.3in Images where LIME-G \underline{underperformed} LIME across all three sensitivity metrics
		\end{flushleft}
	}
	\caption{
		Common images across all three metrics where LIME-G is consistently more robust than LIME (top) and vice versa (bottom).
		Interestingly, we found the intersection of the three sets contains images of mostly scenes, close-up or tiny objects (top).
		In contrast, the common set of images where LIME is more robust than LIME-G contains mostly birds and medium-sized objects (bottom).
	}
	\label{fig:LIME_intersection}
\end{figure}

For each of the three metrics, we computed a set of top-100 score differences between LIME-G vs. LIME.
Interestingly, we found the intersection of the three sets contains images of mostly scenes, close-up or tiny objects (see Fig.~\ref{fig:LIME_intersection}).
In contrast, the common set of images where LIME is more robust than LIME-G contains mostly birds and medium-sized objects.
These image distributions intuitively align with the domains where DeepFill-v1 is capable of inpainting and suggest that the performance of G-methods can be improved further with class-conditional inpainters.

\def\arraystretch{1.2}
\begin{table}[h]
	\caption{
	    The results in this table are the number forms of the ImageNet sensitivity results in Fig.~\ref{fig:sensitivity_score_ImageNet}. 
	    G-methods are more robust to hyperparameters across different sensitivity metrics.
    }
	\label{tab:sensitivity_ImageNet}
	\def\arraystretch{1.5}%
	\begin{center}
	\begin{small}
	\begin{tabular}{|l|c|c|c|}
		\hline
		\multirow{2}{*}{Method} & \multicolumn{3}{c|}{Similarity Metrics}
		\\ \cline{2-4} 
		& SSIM                                            & Pearson correlation of HOGs                        & Spearman               \\ \hline
		SP                   & 0.698$\pm$0.114                 & 0.604$\pm$0.106          & 0.404$\pm$0.261 \\ \hline
		SP-G                   & \textbf{0.781$\pm$0.095} &  \textbf{0.691$\pm$0.093} & 0.317$\pm$0.206 \\ \hline
		LIME \hfill(50)                   & 0.553$\pm$0.060                     & 0.848$\pm$0.028          & 0.573$\pm$0.077 \\ \hline
		LIME-G \hfill(50)                   & \textbf{0.647$\pm$0.057} &  \textbf{0.896$\pm$0.022} & \textbf{0.667$\pm$0.065} \\ \hline
		LIME \hfill(150)                   & 0.163$\pm$0.045                     & 0.708$\pm$0.025          &  0.155$\pm$0.072 \\ \hline
		LIME-G \hfill(150)                   & \textbf{0.371$\pm$0.051} &  \textbf{0.776$\pm$0.022} & \textbf{0.379$\pm$0.059} \\ \hline
		MP2                   & 0.476$\pm$0.155 &  0.453$\pm$0.096 & 0.522$\pm$0.088 \\
		\hline
		MP-G  & \textbf{0.479$\pm$0.064}    & \textbf{0.569$\pm$0.051}          & \textbf{0.698$\pm$0.054}  \\ \hline
	\end{tabular}
	\end{small}
	\end{center}
	\vskip -0.1in
\end{table}

\begin{table}[ht]
	\caption{
	    The results in this table are the number forms of the Places365 sensitivity results in Fig.~\ref{fig:sensitivity_score_Places365}. 
	    The results follow the same trend as the ImageNet dataset.
	}
	\label{tab:sensitivity_Places365}
	\def\arraystretch{1.5}%
	\begin{center}
		\begin{small}
			\begin{tabular}{|l|c|c|c|}
				\hline
				\multirow{2}{*}{Method} & \multicolumn{3}{c|}{Similarity Metrics}                                                                                                                                                                     \\ \cline{2-4} 
				& SSIM                                            & Pearson correlation of HOGs                        & Spearman\\ \hline
				SP & 0.577$\pm$0.177          &  0.674$\pm$0.073          & 0.452$\pm$0.288          \\ \hline
				SP-G  & \textbf{0.720$\pm$0.122} &  \textbf{0.755$\pm$0.056} & \textbf{0.332$\pm$0.208} \\ \hline
				LIME \hfill(50)  & 0.392$\pm$0.074          &  0.802$\pm$0.036          & 0.594$\pm$0.078          \\ \hline
				LIME-G \hfill(50) & \textbf{0.498$\pm$0.076} & \textbf{0.865$\pm$0.027} & \textbf{0.722$\pm$0.058} \\ \hline
				LIME \hfill(150) & 0.118$\pm$0.046          &  0.701$\pm$0.026          & 0.201$\pm$0.071          \\ \hline
				LIME-G  \hfill(150) & \textbf{0.312$\pm$0.061} & \textbf{0.780$\pm$0.022} & \textbf{0.511$\pm$0.051} \\ \hline
				MP2 & 0.466$\pm$0.113 & 0.409$\pm$0.141 & 0.483$\pm$0.140 \\ \hline
				MP2-G & \textbf{0.494$\pm$0.053} & \textbf{0.505$\pm$0.060} & \textbf{0.618$\pm$0.057} \\ \hline
			\end{tabular}
		\end{small}
	\end{center}
	\vskip -0.1in
\end{table}

\begin{figure}[H]
	\centering
	\begin{tabular}{c}
		\includegraphics[width=0.33\linewidth]{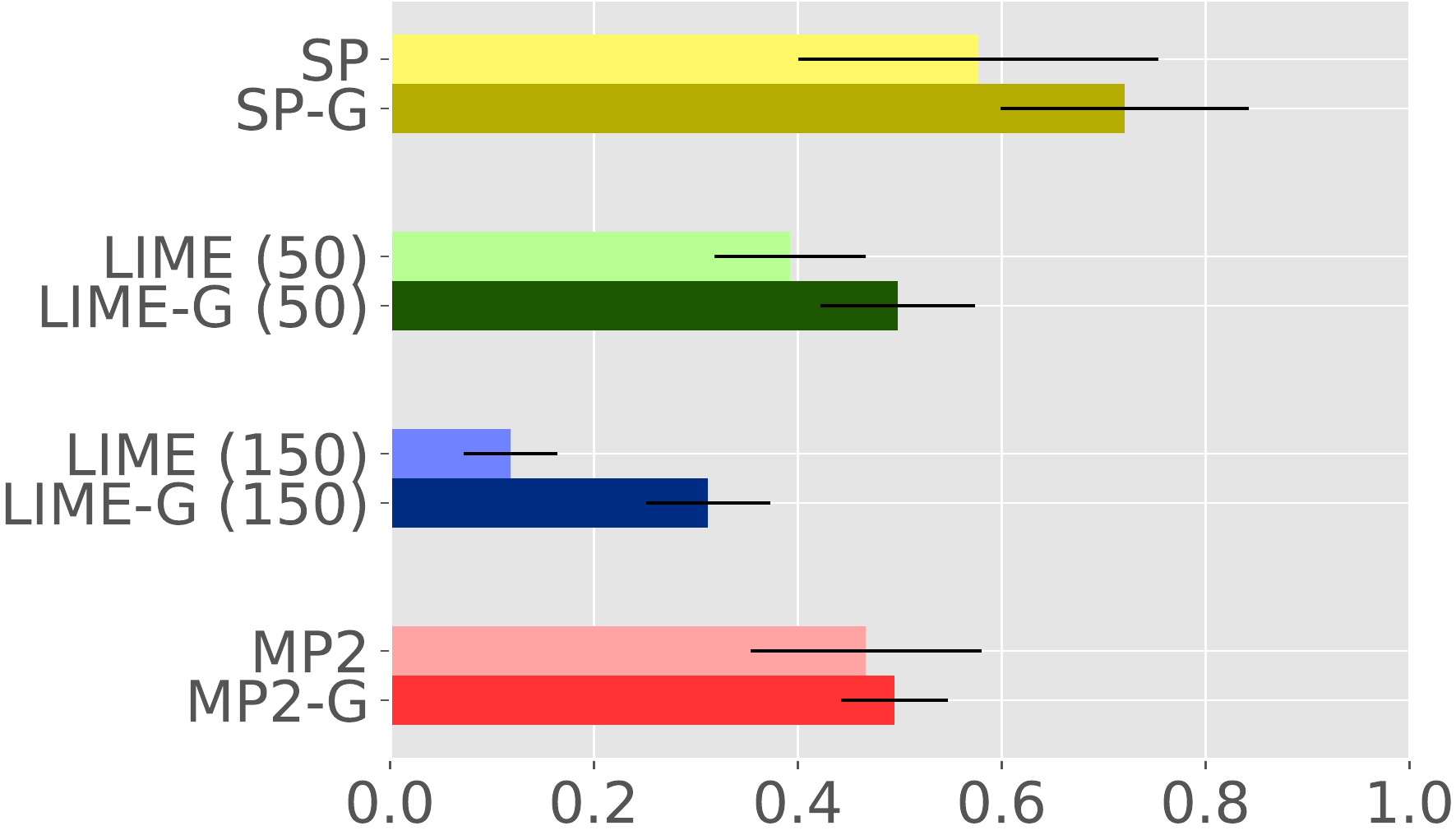}
		\includegraphics[width=0.33\linewidth]{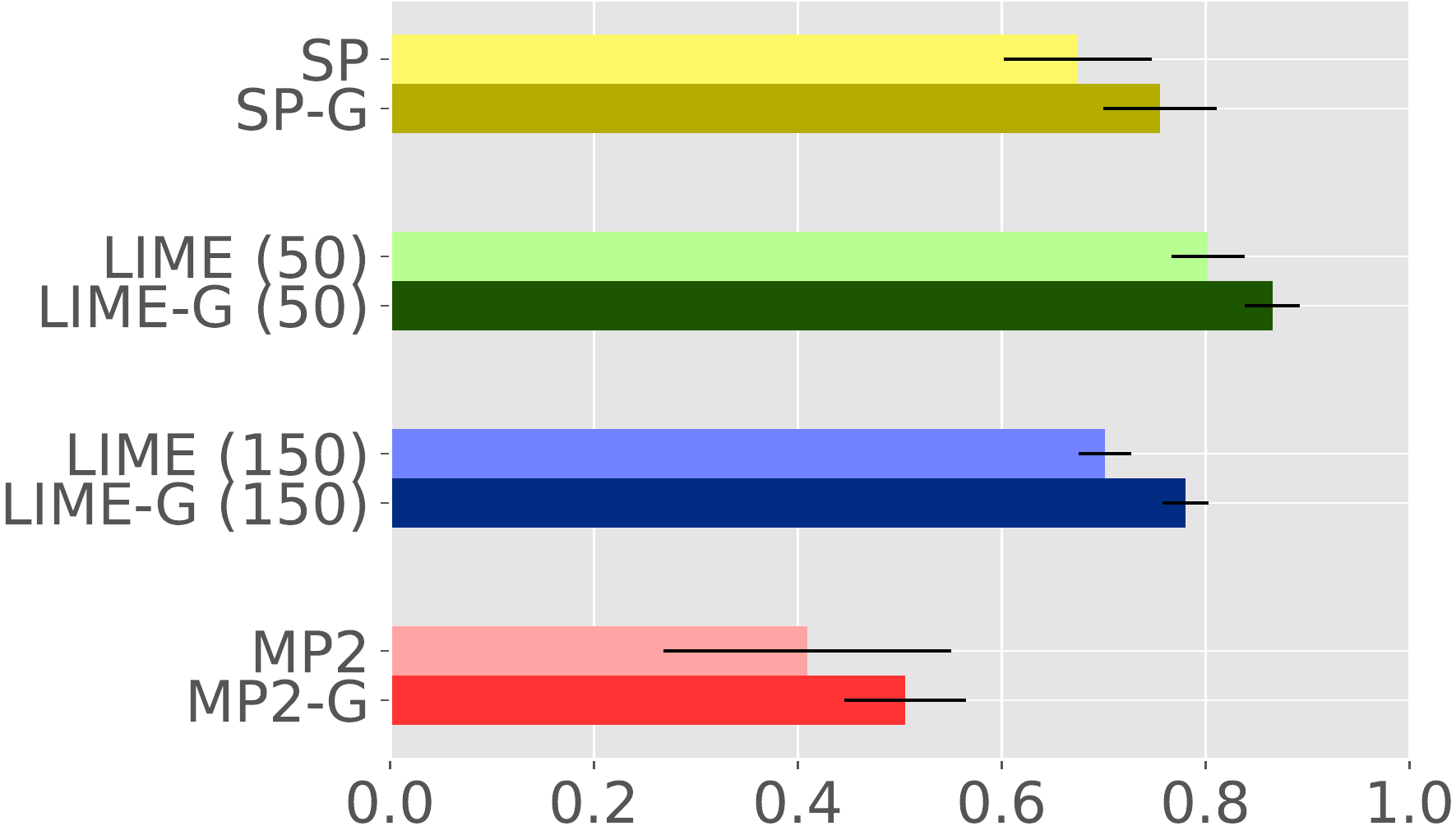} 
		\includegraphics[width=0.33\linewidth]{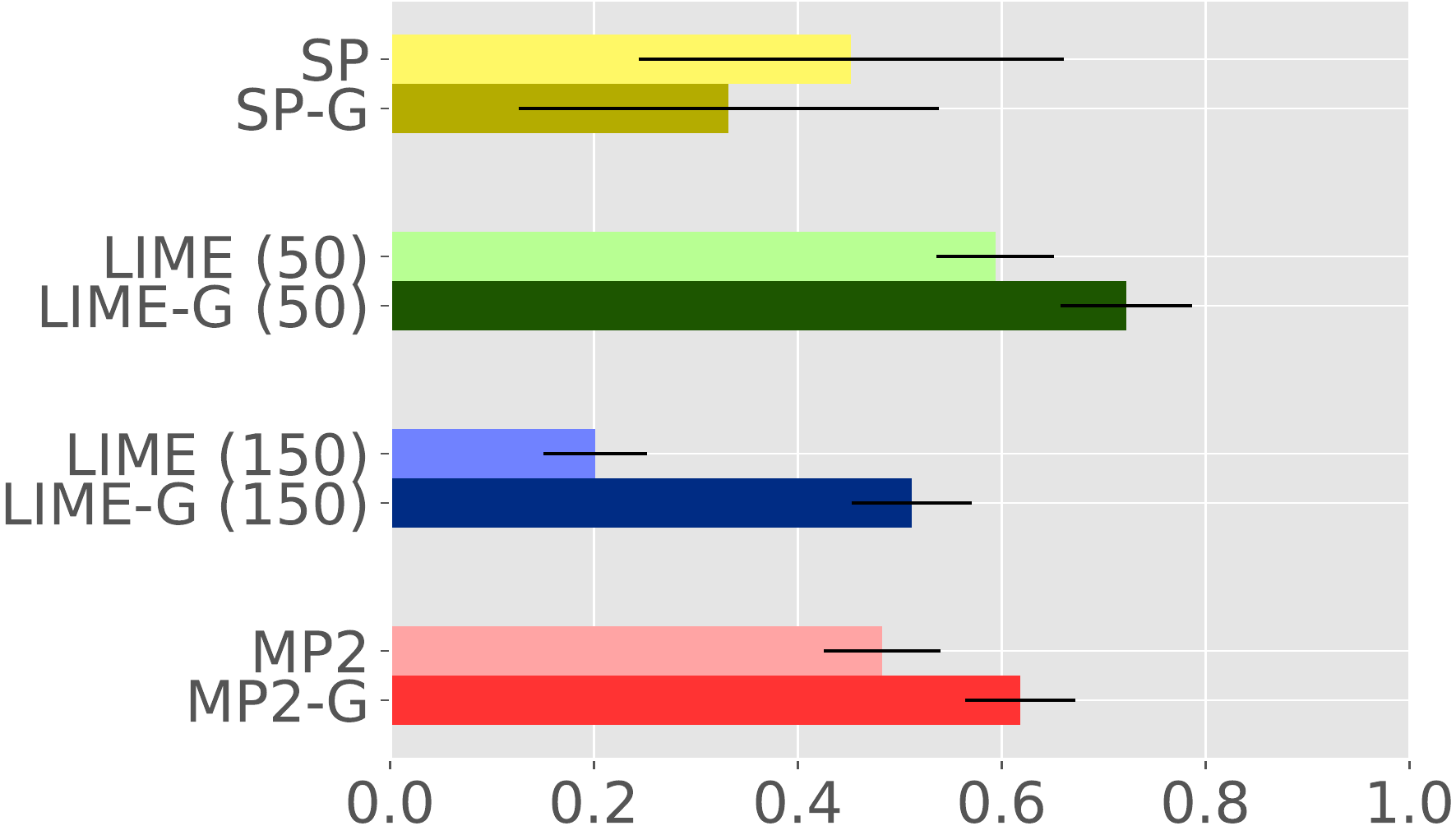} \\
		{	
			\hskip 0.6in (a) SSIM
			\hskip 0.2in (b) Pearson correlation of HOG features
			\hskip 0.02in (c) Spearman rank correlation
		}
	\end{tabular}
	\caption{
		Bar plots comparing the robustness (higher is better) of G-methods and their counterparts when changing hyperparameters (described in Sec.~\ref{sec:sensitivity}) under three different similarity metrics: SSIM (a), Pearson correlation of HOG features (b), 
		and Spearman rank correlation (c).
		Each bar shows the mean and standard deviation similarity score across 1000 pairs of heatmaps, each produced for one random \textbf{Places365} image.
		G-methods are consistently more robust than their counterparts across all metrics.
		The exact numbers are reported in Table~\ref{tab:sensitivity_Places365}.
	}
	\label{fig:sensitivity_score_Places365}
\end{figure}

\begin{figure}[h]
	\centering
	{	
		\begin{flushleft}
			\hskip 0.35in (a) Real (b) Mask (c) Preserve (d) Delete (e) Real (f) Mask (g) Preserve (h) Delete
		\end{flushleft}
	}
	\includegraphics[width=0.85\linewidth]{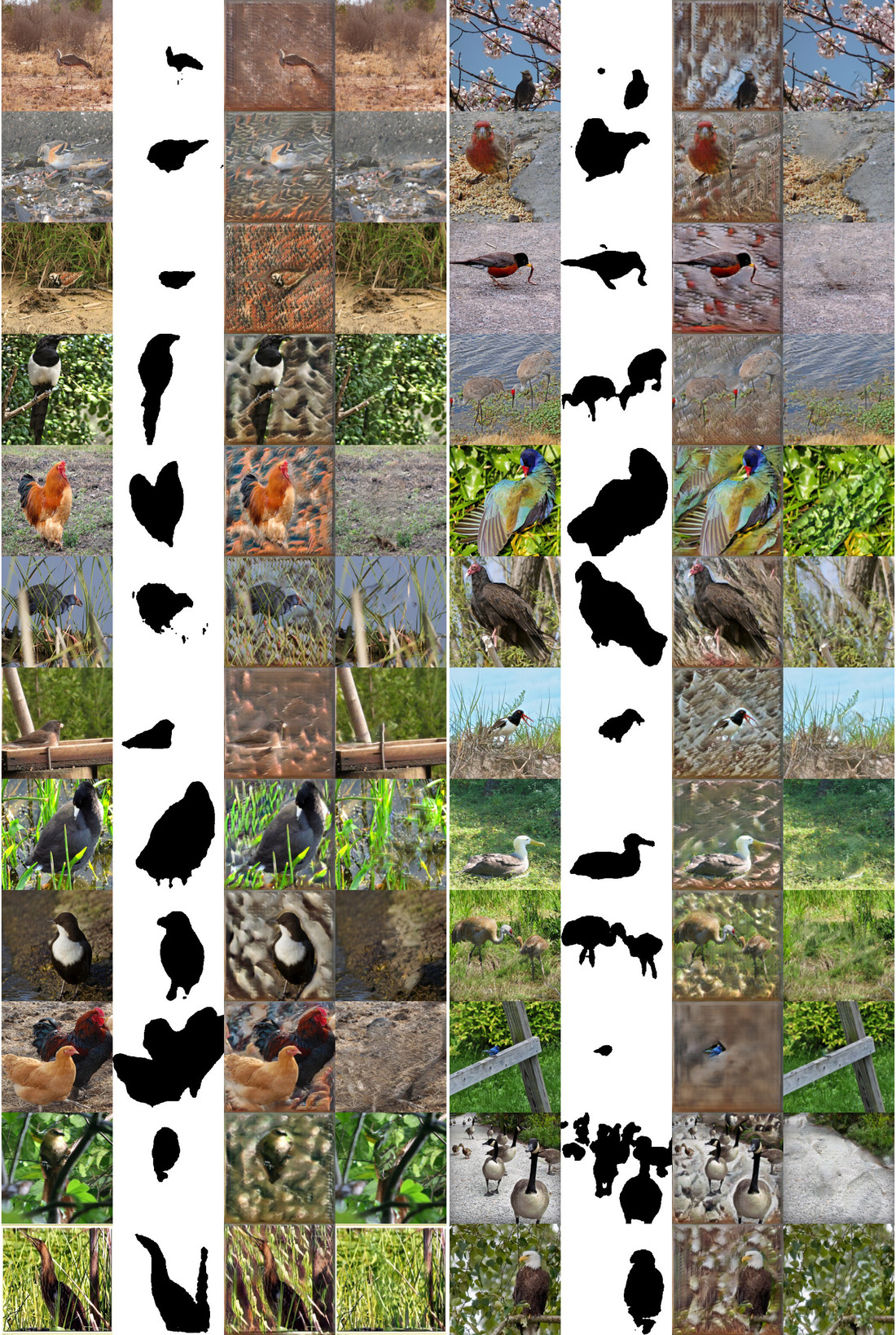}
	\caption{Inpainting using the preservation objective generates unrealistic samples (Sec.\ref{sec:fido}). 
		We randomly chose 50 validation-set images (a) from 52 ImageNet \class{bird} classes and compute their segmentation masks via a pre-trained DeepLab model \cite{chen2017deeplab} (b).
		We found that using the DeepFill-v1 inpainter to inpaint the foreground region (\ie our ``deletion'' task) yields realistic samples where the object is removed (d).
		In contrast, using the inpainter to fill in the background region (\ie ``preservation'' task) yields unrealistic images whose backgrounds contain features (\eg bird feathers or beaks) unnaturally pasted from the object (c).
	}
	\label{fig:FIDO_full}
\end{figure}

\begin{figure}[]
	\centering
	{	
		\footnotesize
		\begin{flushleft}
			\hskip 0.01in LIME-G \underline{outperformed} LIME (top-10 cases)
			\hskip 0.1in LIME-G \underline{underperformed} LIME (top-10 cases)
		\end{flushleft}
	}
	{	
		\begin{flushleft}
			\hskip 0.12in Real
			\hskip 0.16in LIME
			\hskip 0.52in LIME-G
			\hskip 0.65in Real
			\hskip 0.17in LIME
			\hskip 0.52in LIME-G
		\end{flushleft}		
	}
	\includegraphics[width=1.0\linewidth]{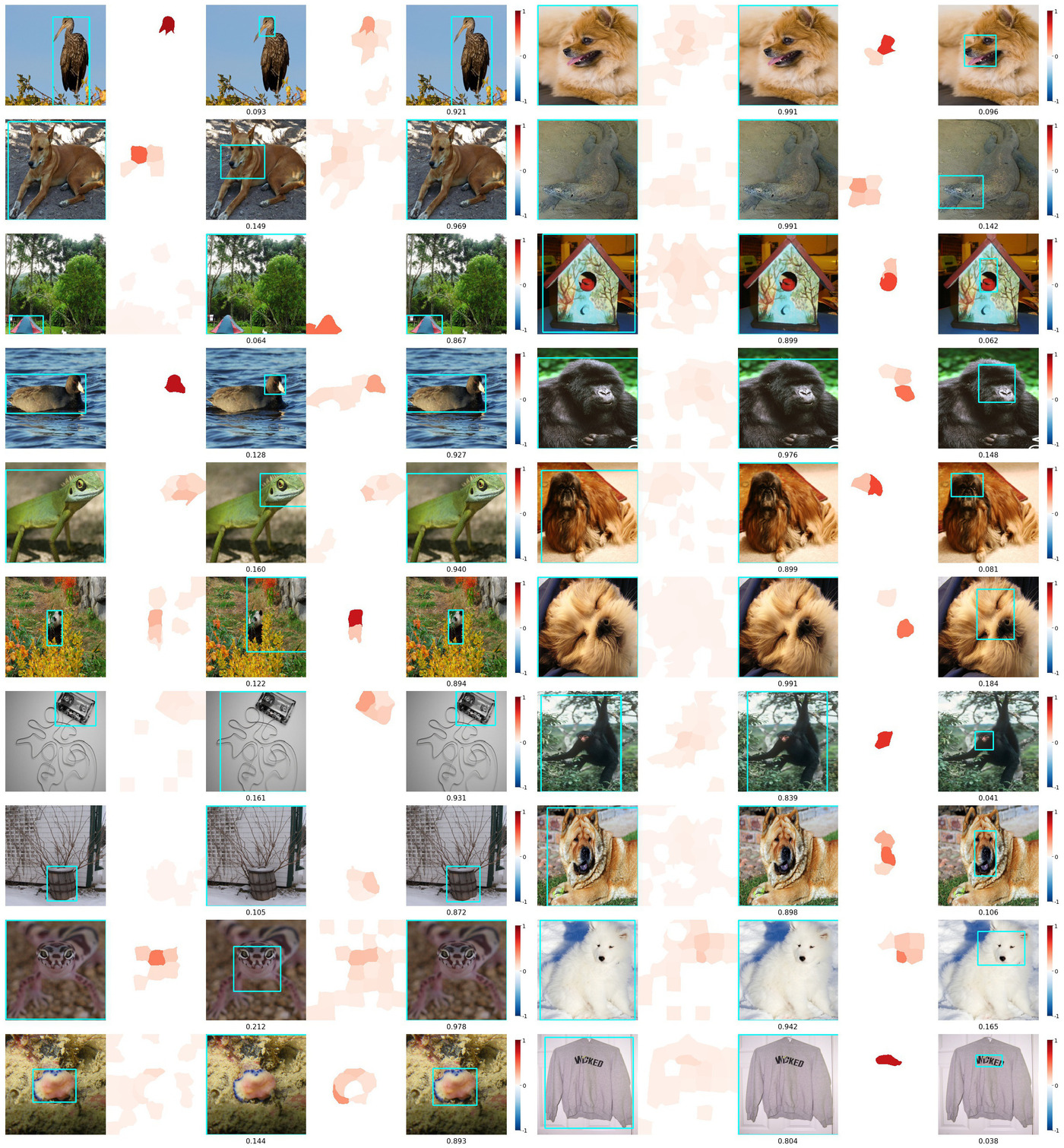}
	\caption{Top-10 cases where the LIME-G outperformed (left) and underperformed (right) LIME on the object localization task (IoU scores).
		From left to right, on each row: we show a real image with its ground-truth bounding box, LIME heatmap \& its derived bounding box, LIME-G heatmap \& its derived bounding box.
		See \url{https://drive.google.com/drive/u/2/folders/10JeP9dpuoa0M16xe2FloBEWajQ7PNKSX} for more examples of the LIME and LIME-G IoU results.}
	\label{fig:IOU_LIME}
\end{figure}

\begin{figure}[]
	\vskip 0.1in
	\centering
	{	
		\begin{flushleft}
			\hskip 0.12in Top-10 cases where SP-G \underline{outperformed} SP
			\hskip 0.15in Top-10 cases where SP-G \underline{underperformed} SP
		\end{flushleft}
	}
	{	
		\begin{flushleft}
			\hskip 0.12in Real
			\hskip 0.26in SP
			\hskip 0.65in SP-G
			\hskip 0.75in Real
			\hskip 0.26in SP
			\hskip 0.66in SP-G
		\end{flushleft}		
	}
	\includegraphics[width=1.0\linewidth]{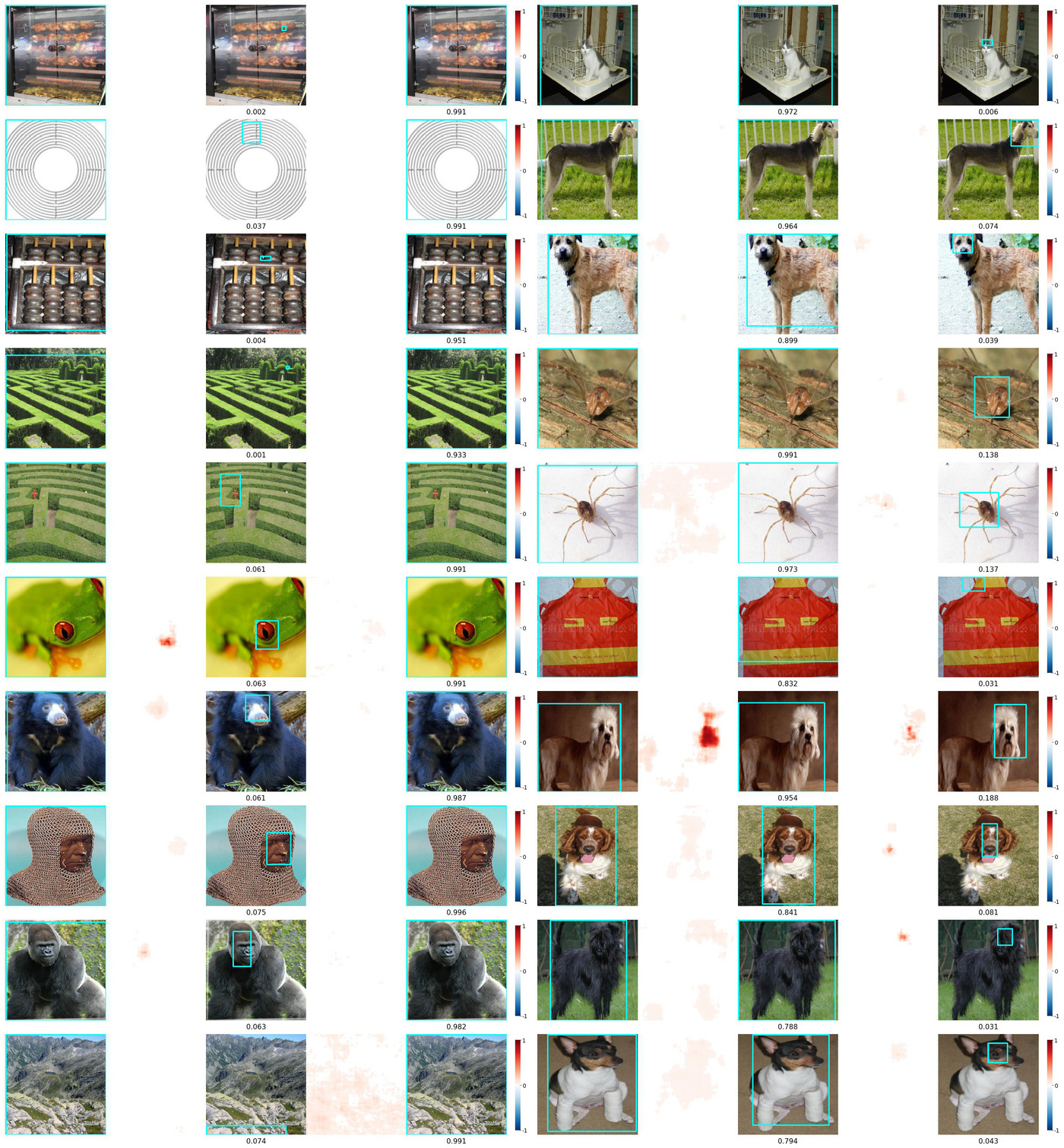}
	\caption{Top-10 cases where the SP-G outperformed (left) and underperformed (right) SP on the object localization task (IoU scores).  
		From left to right, on each row: we show a real image with its ground-truth bounding box, SP heatmap \& its derived bounding box, SP-G heatmap \& its derived bounding box.
		In the cases where SP-G has a lower IoU score than SP (right panel), we observed the heatmap localizes some unique features of the object as compared to the images in the top cases where the heatmap covers the entire image. 
		See \url{https://drive.google.com/drive/u/2/folders/1XJ6M0AMHxZrXxLLw6m3Bx7sjvsyqN6JC} for more examples of the SP and SP-G IoU results.}
	\label{fig:IOU_occlusion}
\end{figure}

\begin{figure}[h]
	\subcaptionbox{$\alpha$ vs Localization error for ImageNet \label{fig:IOU_thresh}}%
	[0.99\linewidth]{\includegraphics[width=0.79\columnwidth]{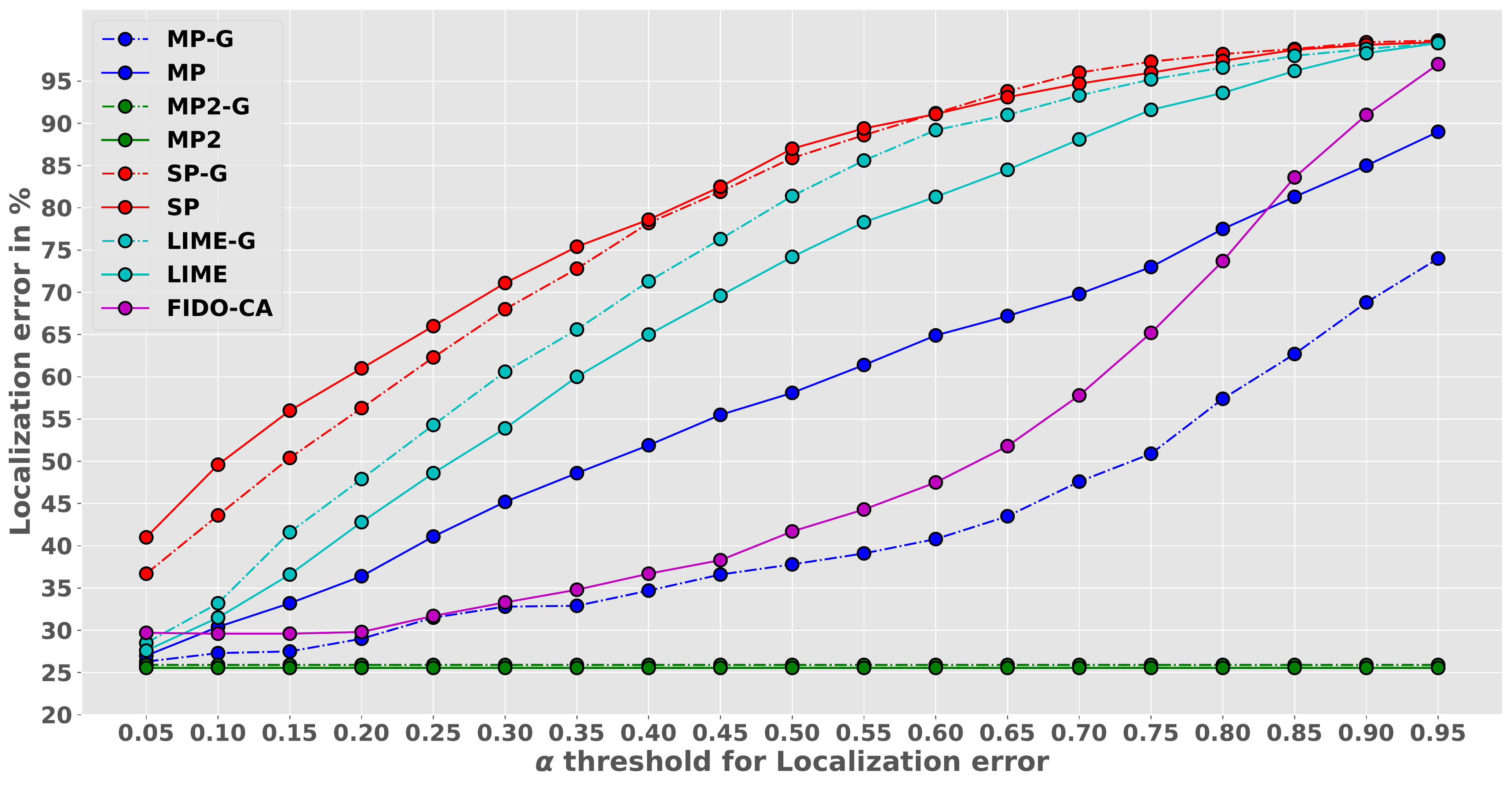}}
	\subcaptionbox{$\alpha$ vs Saliency Metric for ImageNet \label{fig:SM_imagenet_thresh}}%
	[0.99\linewidth]{\includegraphics[width=0.79\columnwidth]{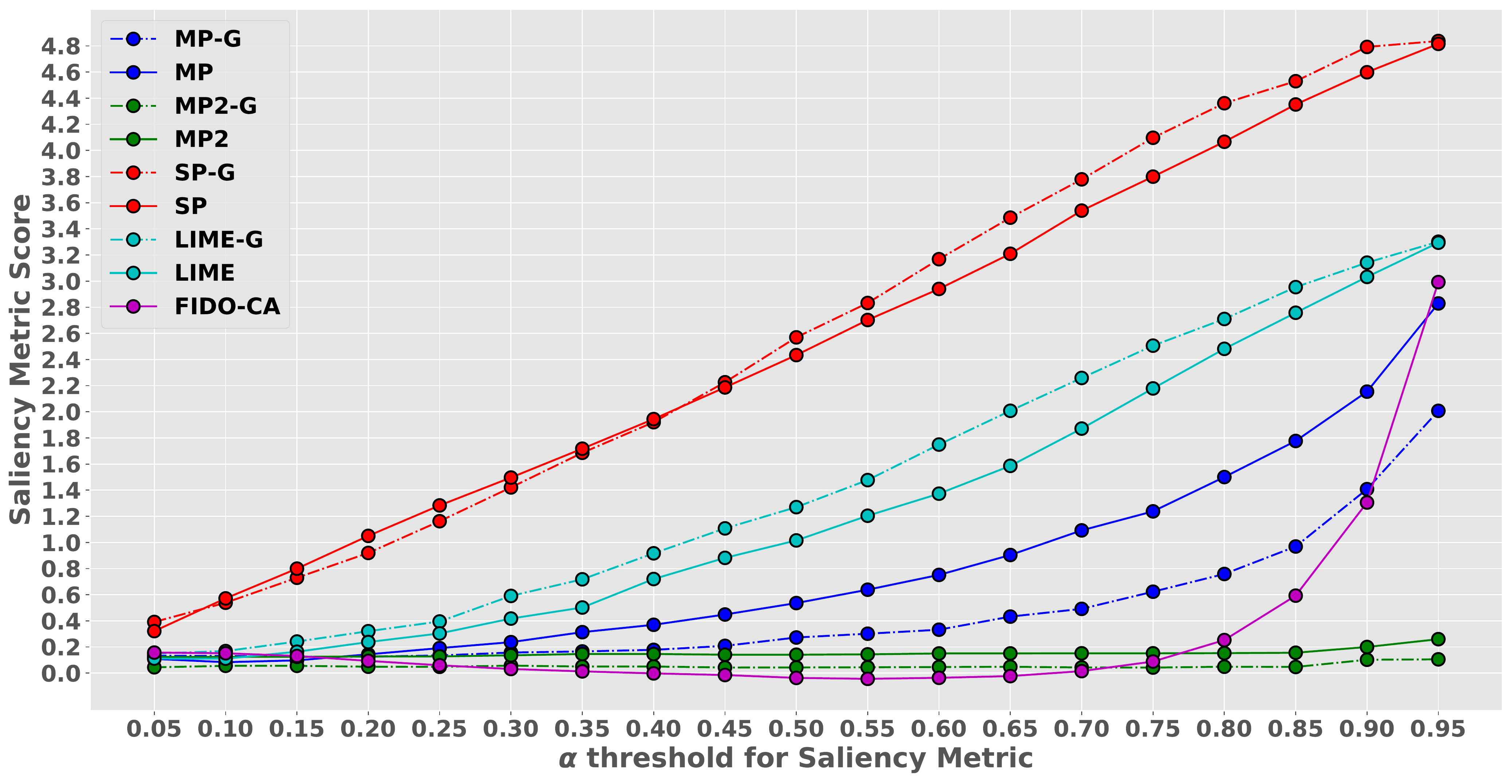}}
	\subcaptionbox{$\alpha$ vs Saliency Metric for Places365 \label{fig:SM_places_thresh}}%
	[0.99\linewidth]{\includegraphics[width=0.79\columnwidth]{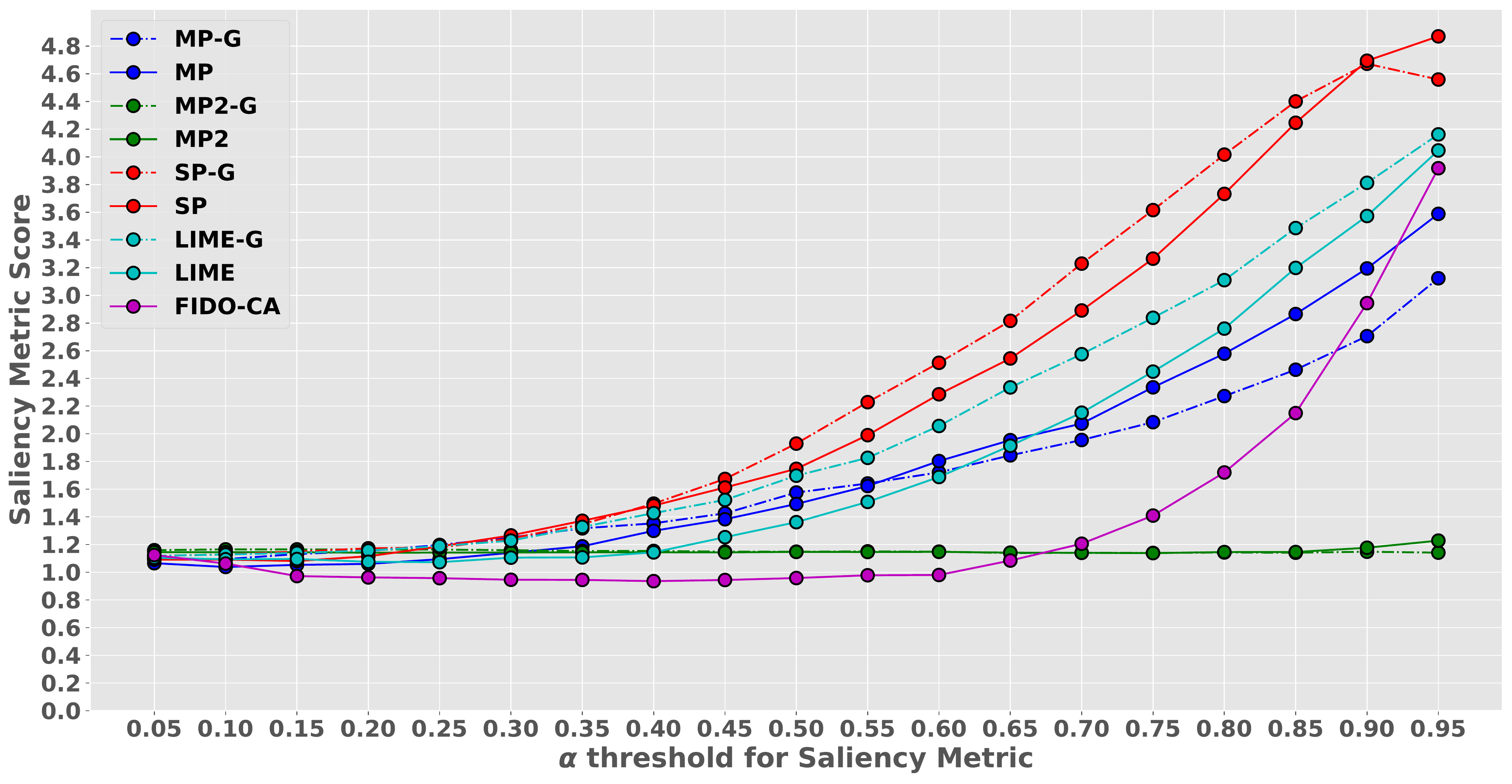}}
	\caption{
	    Localization error (a) and saliency metric (b, c) performance of different attribution methods on a held-out set of 1000 images for different $\alpha$ threshold values.
	    For each method, we search for the optimal $\alpha$ value on this held-out set and use the subsequent threshold for computing the scores on the 2000 images in the object localization and saliency metric experiments in Sec.~\ref{sec:faithful}.
	}
	\label{fig:alpha_performance}
\end{figure}

\begin{figure}[]
	\includegraphics[width=1.0\linewidth]{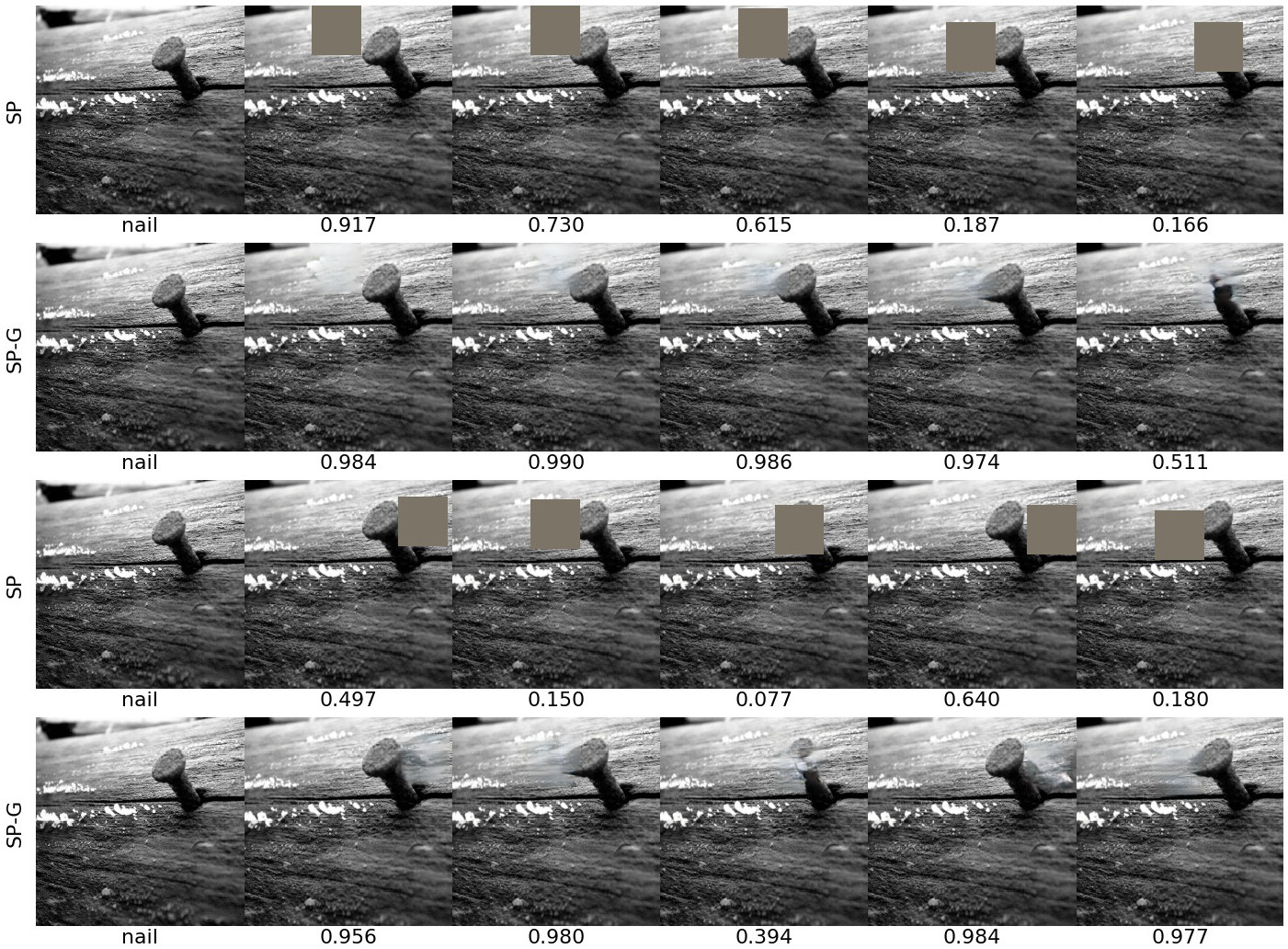}
	\caption{
		Random intermediate perturbation samples by SP and SP-G on the same image from the \class{nail} class in ImageNet.
		SP-G drops the target-class probability only when the patch cover a major area of the nail (e.g. the center $0.394$-probability sample in the bottom panel). 
		This figure is a zoom-in version of the samples in Fig.~\ref{fig:SP_teaser}. 
	}
	\label{fig:SP_teaser_perturbed}
\end{figure}

\begin{figure}[h]
	\centering
	\includegraphics[width=1.0\linewidth]{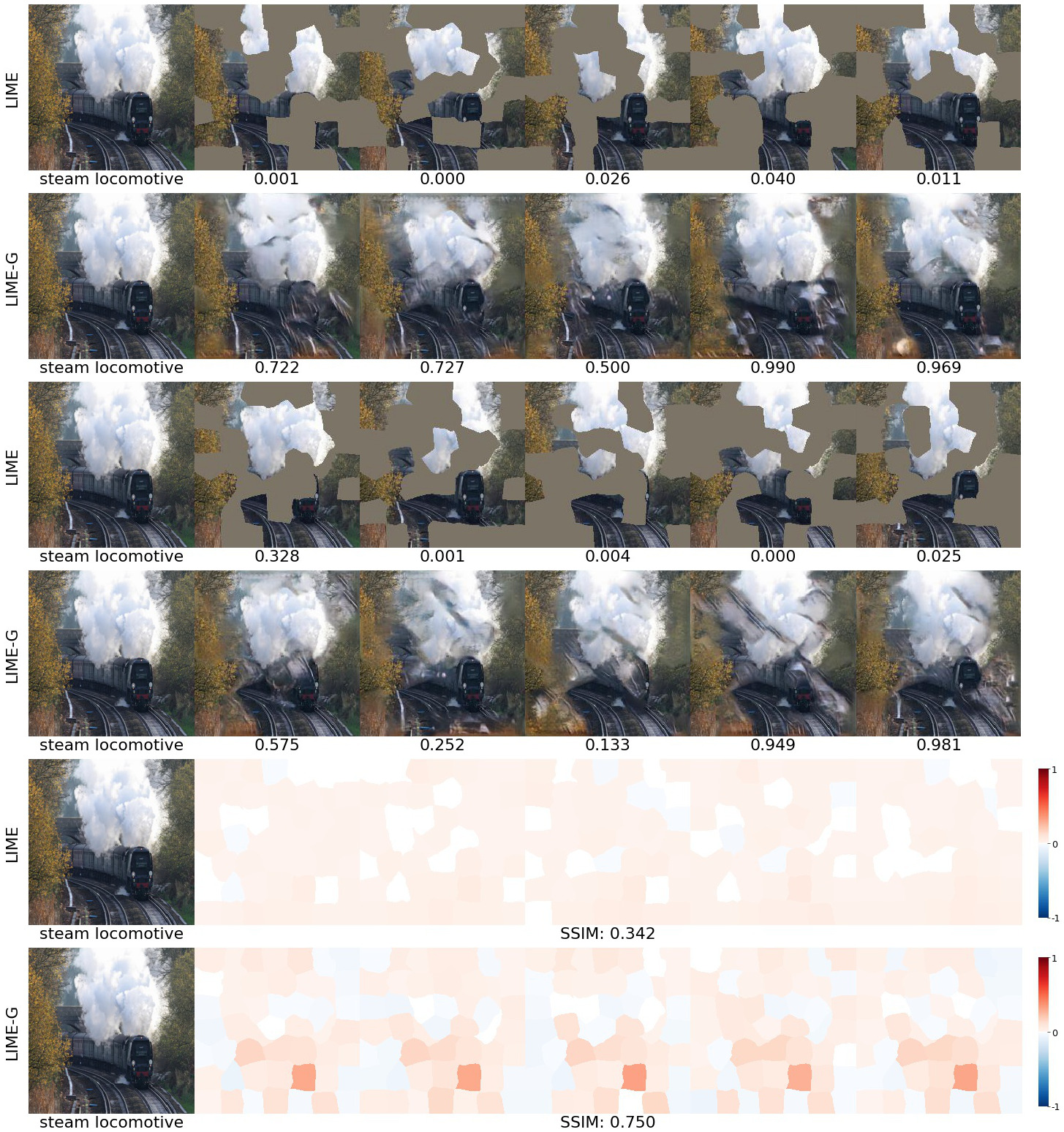}
	\caption{
		Qualitative evidence supporting the LIME-G vs. LIME sensitivity experiment in Sec.~\ref{sec:sensitivity}.
		For both LIME and LIME-G, per image, we compute an average SSIM score across all 10 pairs of 5 heatmaps.
		We then take the difference between LIME-G and LIME and sort them in the descending order.
		This \class{steam~locomotive} image is a random image from the top-100 ImageNet-S cases where LIME-G \underline{outperformed} LIME.
		\textbf{Top four rows:} Here, we compare pairs of LIME vs. LIME-G perturbation samples that were created from the same random superpixel masks. 
		LIME-G samples cause large probability drops only when some discriminative feature is removed from the image and thus results in more localized heatmaps. 
		\textbf{Bottom two rows:} 5 heatmaps by LIME and LIME-G, each from a random seed.
		While LIME-G heatmaps are more consistent, LIME heatmaps is noisy and varies.
		See Fig.~\ref{fig:LIME_ImageNet_top_sensitivity_2} and Figs.~\ref{fig:LIME_Places365_top_sensitivity_1}-\ref{fig:LIME_Places365_top_sensitivity_2} for similar observations in ImageNet-S and Places365-S dataset respectively. See \url{https://drive.google.com/drive/u/2/folders/1sKWig4Xk5Pm50kdONdAS9SkiTBhJRAkw} for more examples.}
	\label{fig:LIME_ImageNet_top_sensitivity_1}
\end{figure}

\begin{figure}[h]
	\centering
	\includegraphics[width=1.0\linewidth]{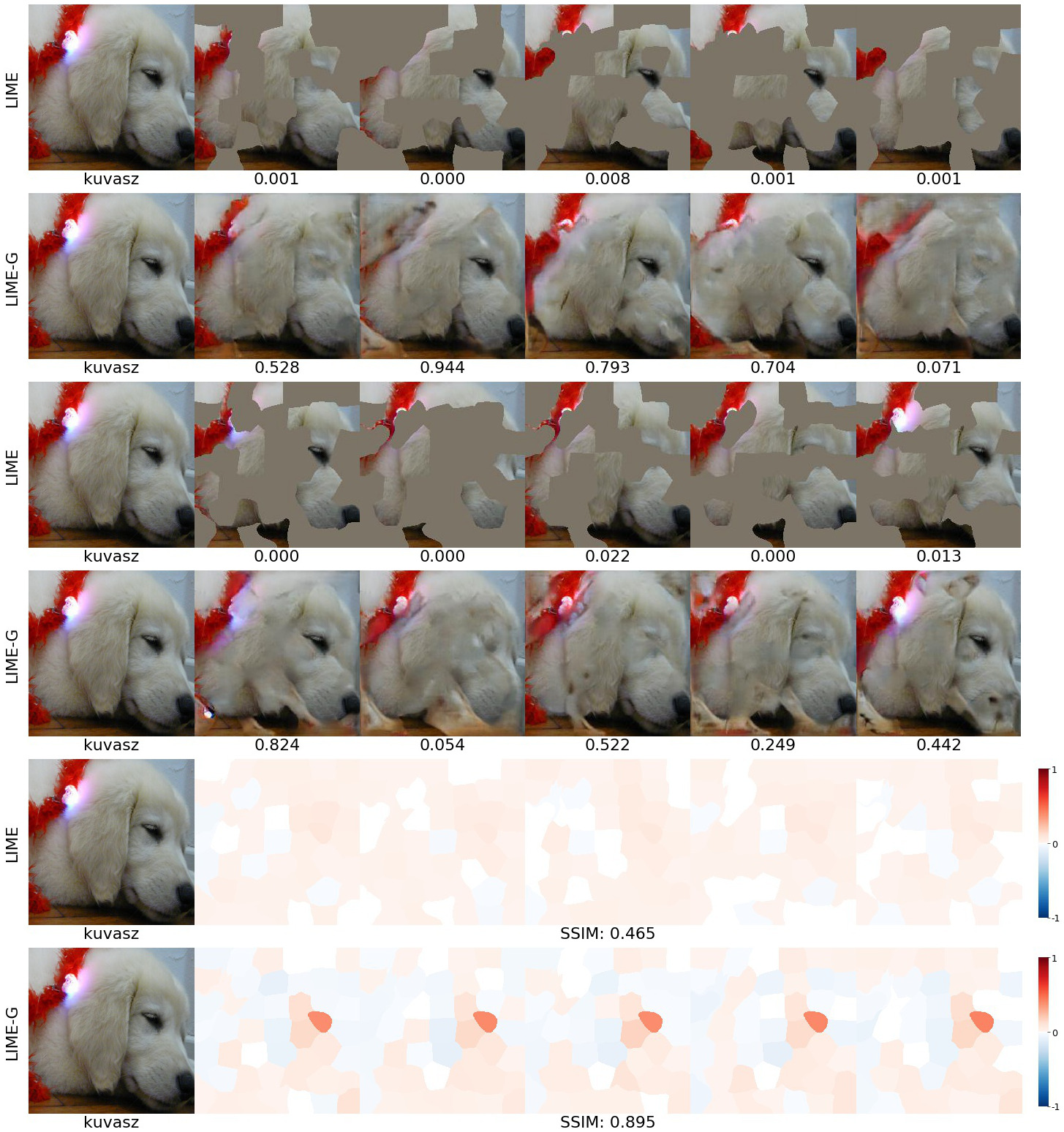}
	\caption{
		Here, we show the same figure as Fig.~\ref{fig:LIME_ImageNet_top_sensitivity_1} (see its caption) but for another random image among the top-100 ImageNet-S cases where LIME-G \underline{outperformed} LIME on the SSIM similarity metric.
		See \url{https://drive.google.com/drive/u/2/folders/1sKWig4Xk5Pm50kdONdAS9SkiTBhJRAkw} for more examples.}
	\label{fig:LIME_ImageNet_top_sensitivity_2}
\end{figure}

\begin{figure}[]
	\centering
	\includegraphics[width=1.0\linewidth]{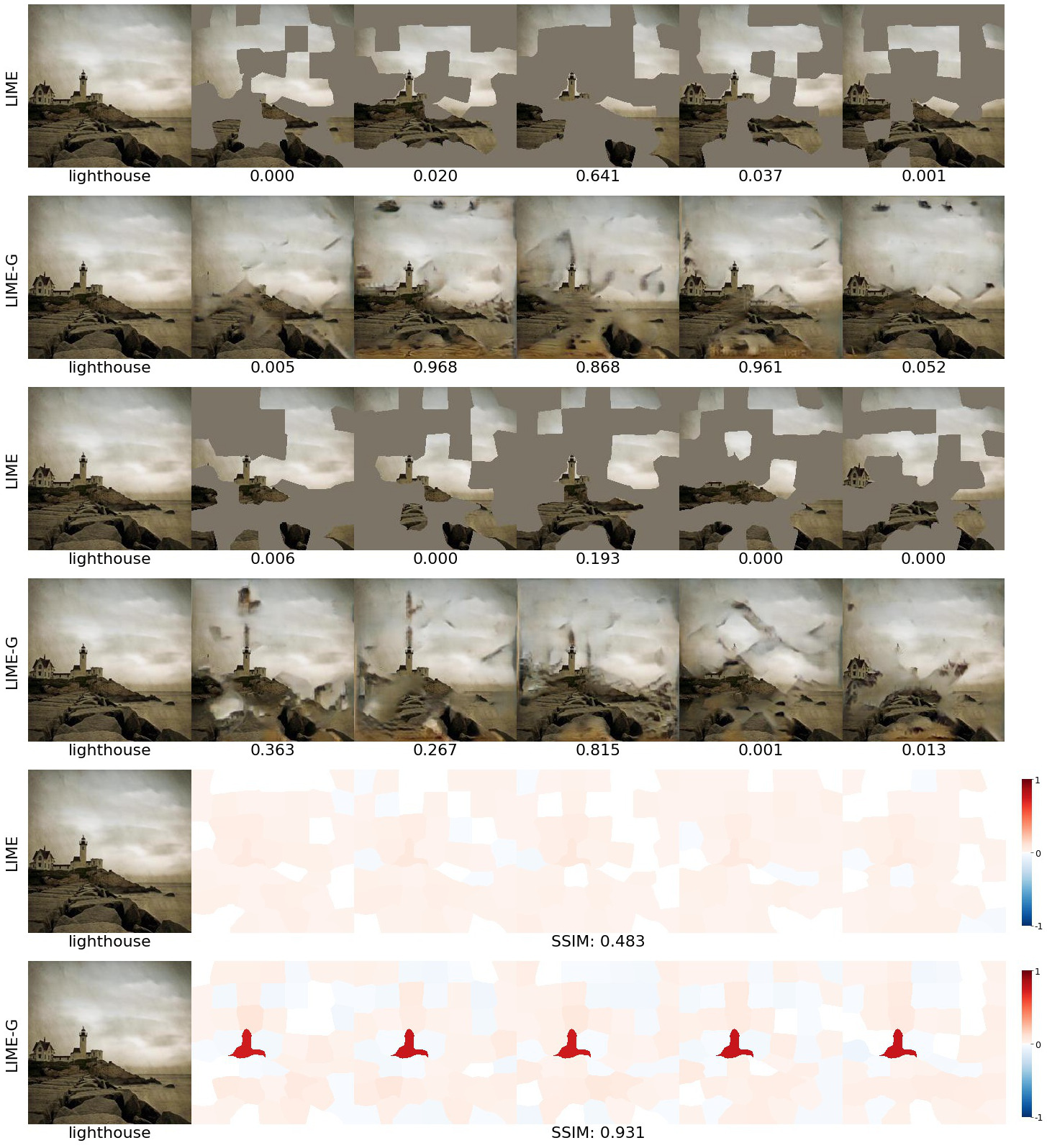}
	\caption{
		Here, we show the same figure as Fig.~\ref{fig:LIME_ImageNet_top_sensitivity_1} (see its caption) but for a random image among the top-100 Places365-S cases where LIME-G \underline{outperformed} LIME on the SSIM similarity metric.
		See \url{https://drive.google.com/drive/u/2/folders/1aXyDFBq0HlcI0kQJpJyspNf2rtwLj35Z} for more examples.}
	\label{fig:LIME_Places365_top_sensitivity_1}
\end{figure}

\begin{figure}[]
	\centering
	\includegraphics[width=1.0\linewidth]{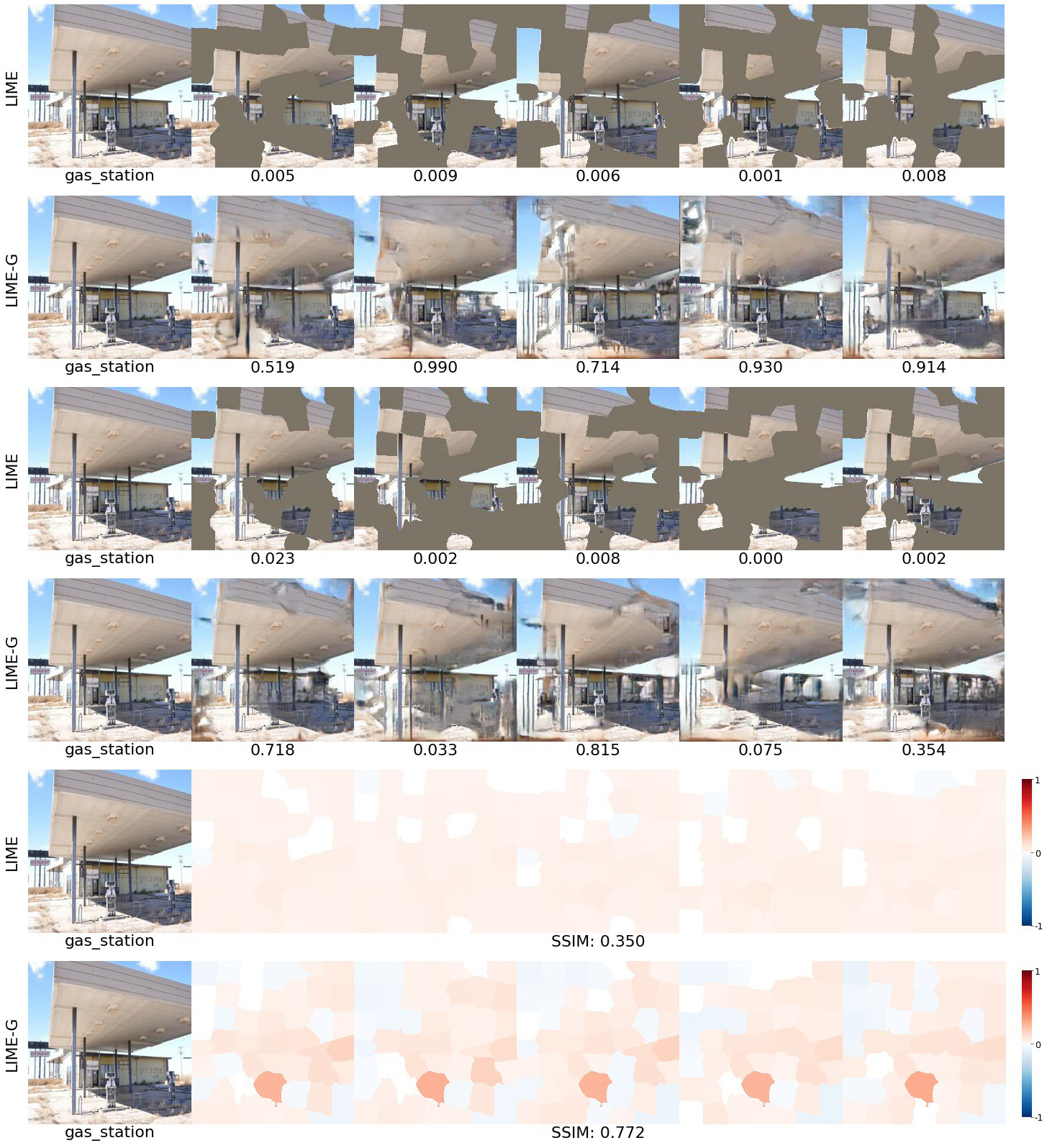}
	\caption{
		Here, we show the same figure as Fig.~\ref{fig:LIME_ImageNet_top_sensitivity_1} (see its caption) but for a random image among the top-100 Places365-S cases where LIME-G \underline{outperformed} LIME on the SSIM similarity metric.
		See \url{https://drive.google.com/drive/u/2/folders/1aXyDFBq0HlcI0kQJpJyspNf2rtwLj35Z} for more examples.}
	\label{fig:LIME_Places365_top_sensitivity_2}
\end{figure}

\begin{figure}[h]
	\centering
	\includegraphics[width=1.0\linewidth]{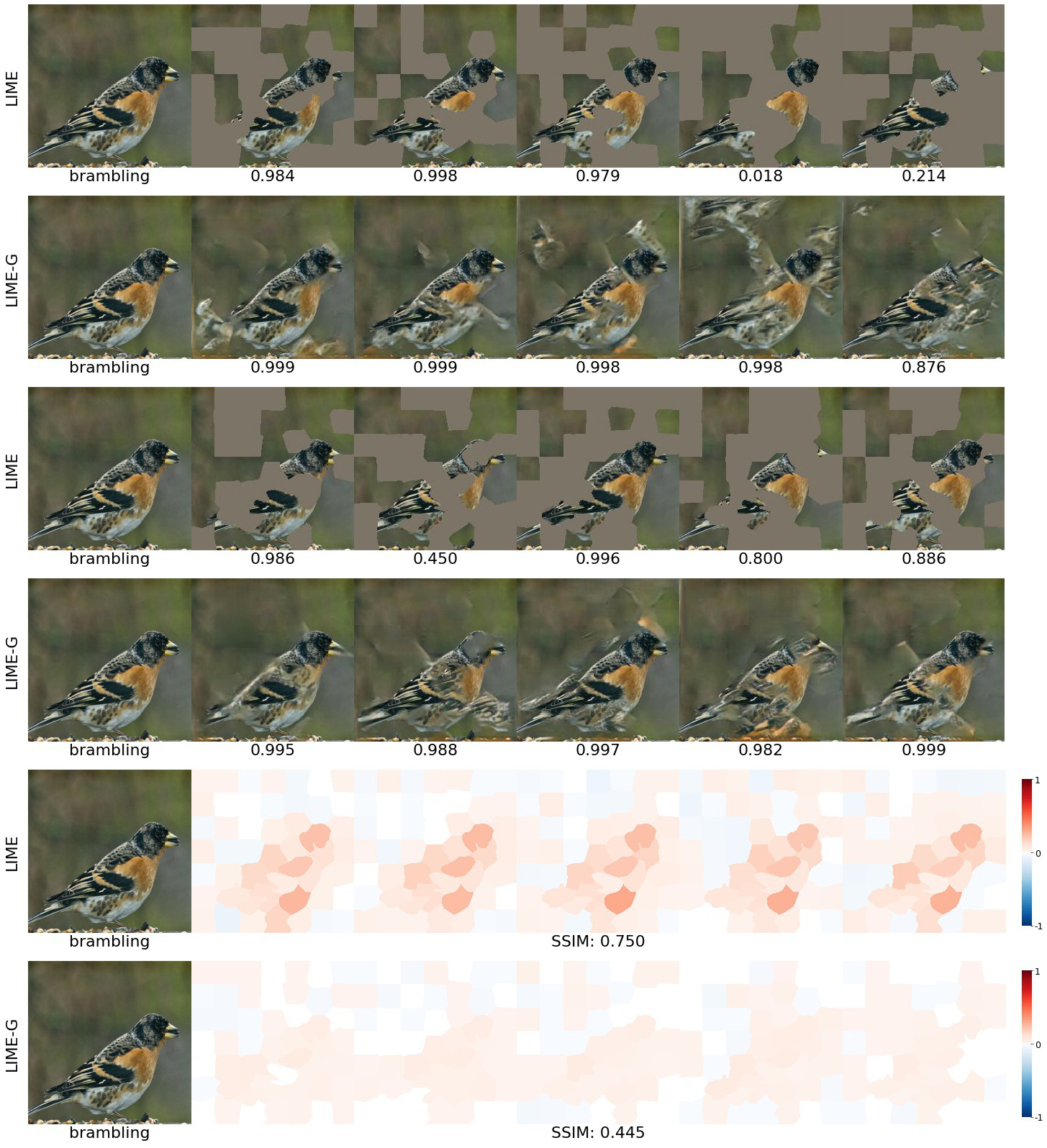}
	\caption{
		Here, we show the same figure as Fig.~\ref{fig:LIME_ImageNet_top_sensitivity_1} (see its caption) but for a random image among the top-100 ImageNet-S cases where LIME-G \underline{underperformed} LIME on the SSIM similarity metric.
		LIME-G samples remain at high target-class probabilities and therefore produced heatmaps that are more sensitive than those of LIME.
		Similar observations can be found in Fig.~\ref{fig:LIME_ImageNet_bottom_sensitivity_2} and Figs.~\ref{fig:LIME_Places365_bottom_sensitivity_1}-\ref{fig:LIME_Places365_bottom_sensitivity_2}. 
		See \url{https://drive.google.com/drive/u/2/folders/1sKWig4Xk5Pm50kdONdAS9SkiTBhJRAkw} for more examples.}
	\label{fig:LIME_ImageNet_bottom_sensitivity_1}
\end{figure}

\begin{figure}[h]
	\centering
	\includegraphics[width=1.0\linewidth]{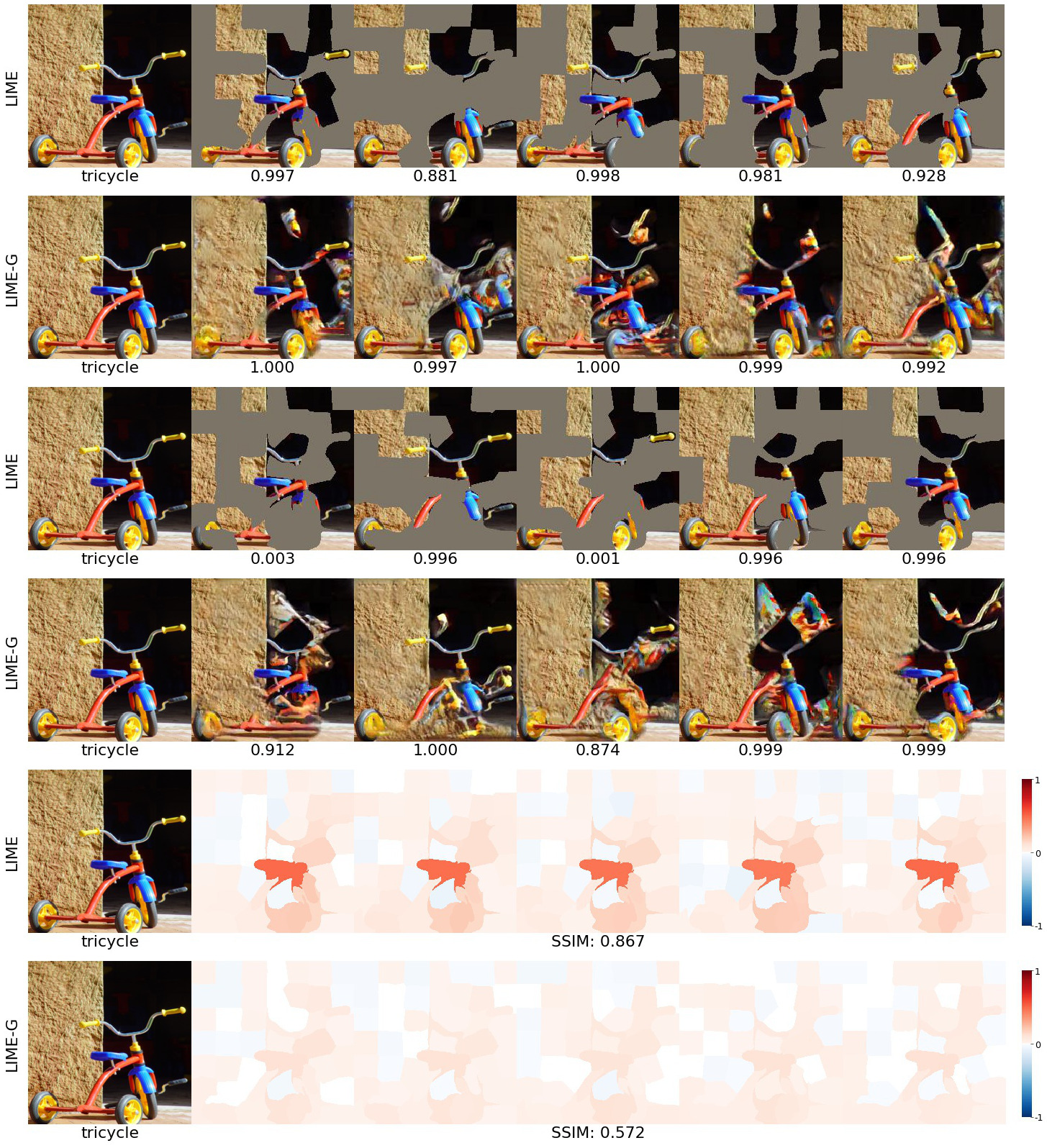}
	\caption{
		Here, we show the same figure as Fig.~\ref{fig:LIME_ImageNet_top_sensitivity_1} (see its caption) but for a random image among the top-100 ImageNet-S cases where LIME-G \underline{underperformed} LIME on the SSIM similarity metric.
		See \url{https://drive.google.com/drive/u/2/folders/1sKWig4Xk5Pm50kdONdAS9SkiTBhJRAkw} for more examples.}
	\label{fig:LIME_ImageNet_bottom_sensitivity_2}
\end{figure}

\begin{figure}[]
	\centering
	\includegraphics[width=1.0\linewidth]{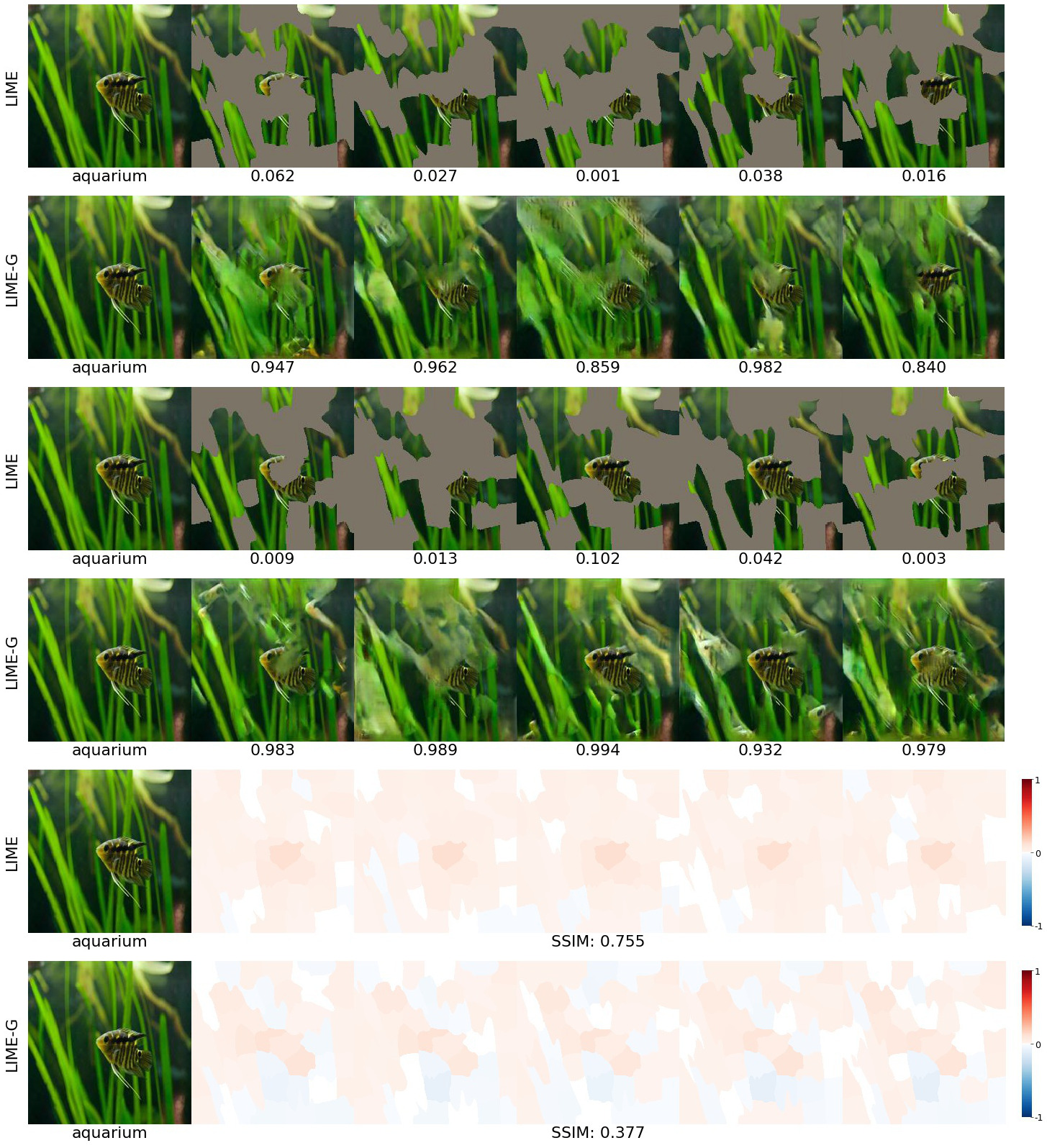}
	\caption{
		Here, we show the same figure as Fig.~\ref{fig:LIME_ImageNet_top_sensitivity_1} (see its caption) but for a random image among the top-100 Places365-S cases where LIME-G \underline{underperformed} LIME on the SSIM similarity metric.
		See \url{https://drive.google.com/drive/u/2/folders/1aXyDFBq0HlcI0kQJpJyspNf2rtwLj35Z} for more examples.}
	\label{fig:LIME_Places365_bottom_sensitivity_1}
\end{figure}

\begin{figure}[]
	\centering
	\includegraphics[width=1.0\linewidth]{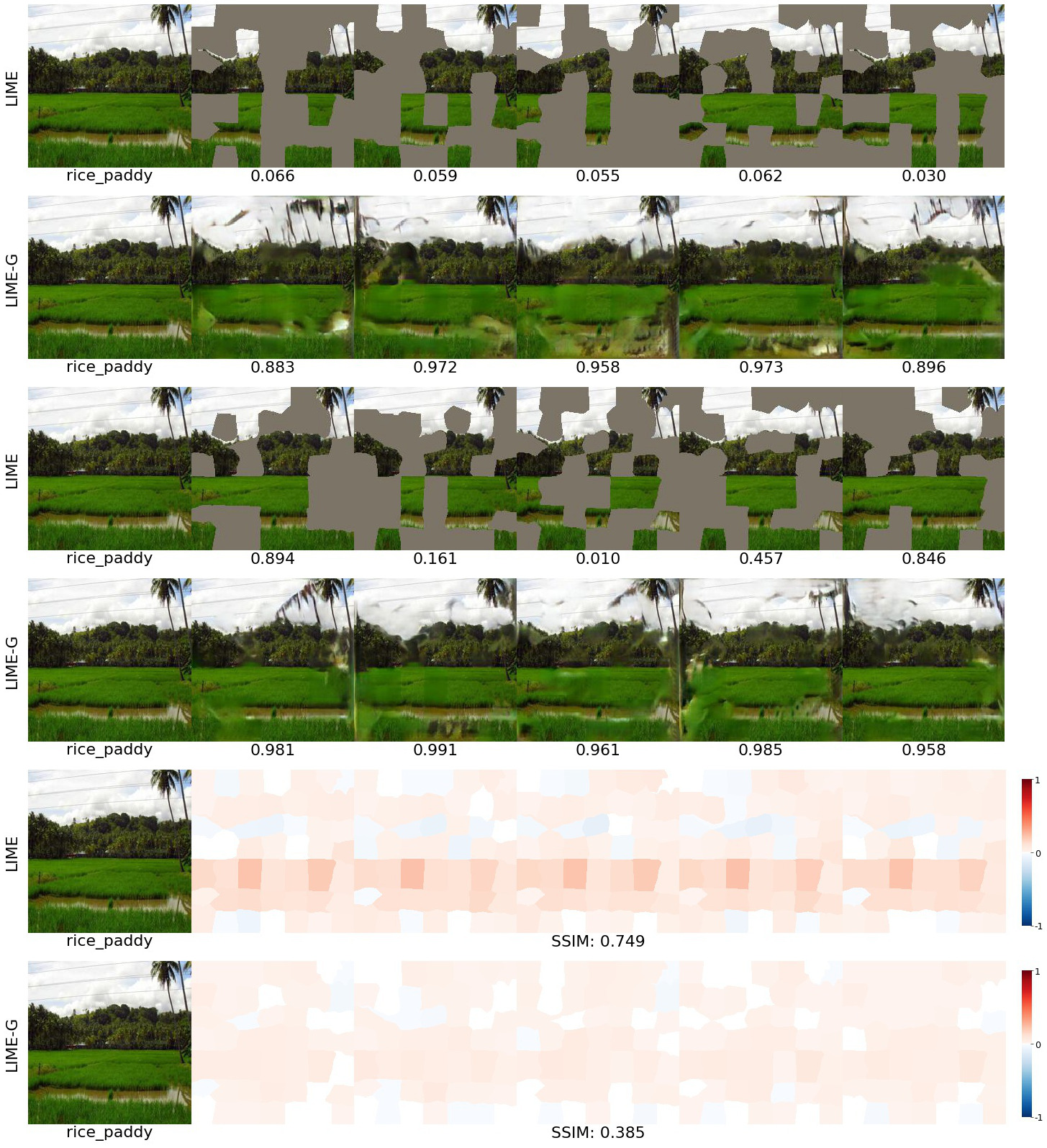}
	\caption{
		Here, we show the same figure as Fig.~\ref{fig:LIME_ImageNet_top_sensitivity_1} (see its caption) but for a random image among the top-100 Places365-S cases where LIME-G \underline{underperformed} LIME on the SSIM similarity metric.
		See \url{https://drive.google.com/drive/u/2/folders/1aXyDFBq0HlcI0kQJpJyspNf2rtwLj35Z} for more examples.}
	\label{fig:LIME_Places365_bottom_sensitivity_2}
\end{figure}

\begin{figure}
	\centering
	\subcaptionbox{LIME histogram distribution is skewed\label{fig:LIME_hist_full}}%
	[.99\textwidth]{\includegraphics[width=\linewidth]{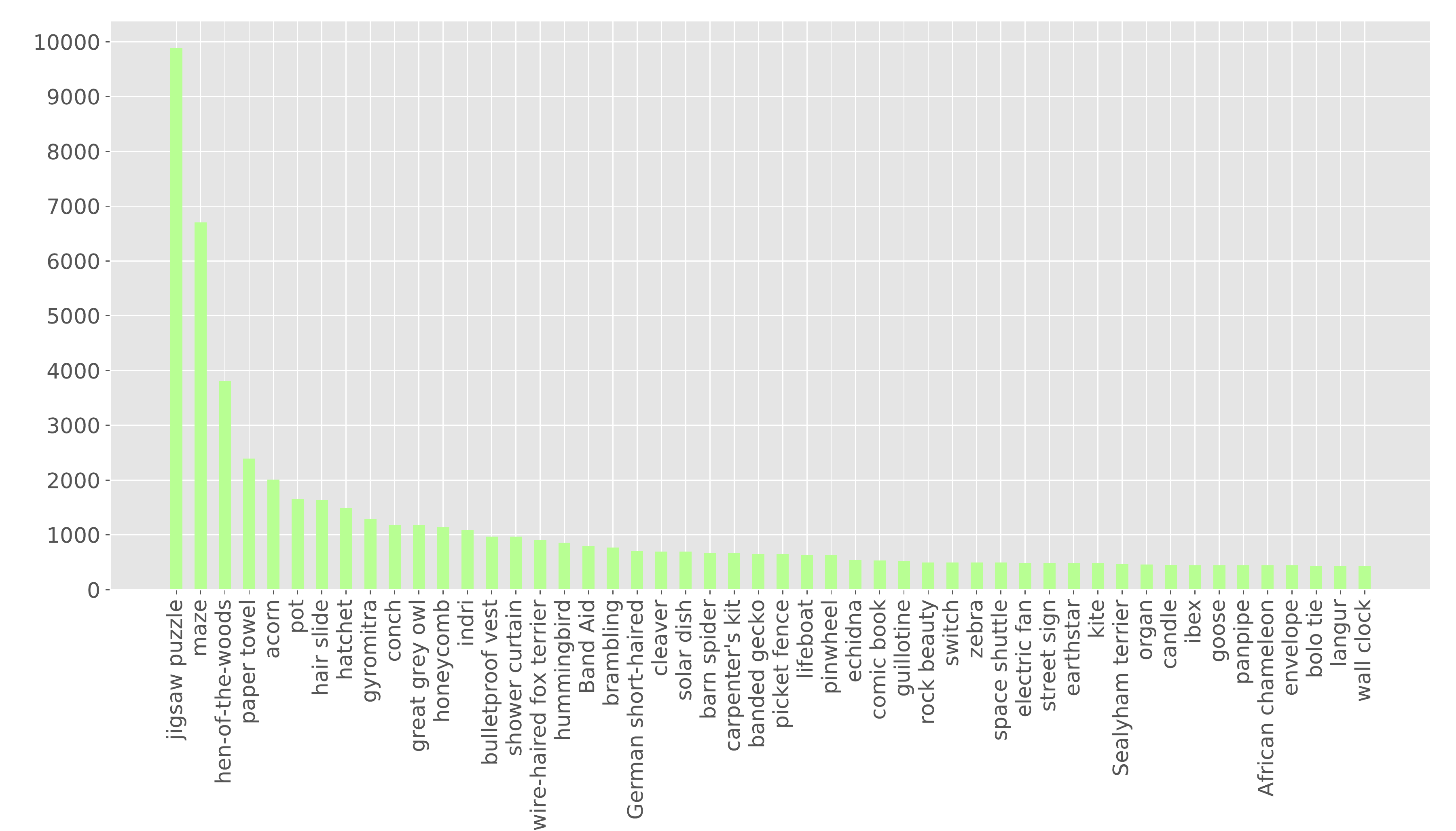}}\\
	\subcaptionbox{LIME-G histogram distribution is almost uniform\label{fig:LIMEG_hist_full}}%
	[.99\textwidth]{\includegraphics[width=\linewidth]{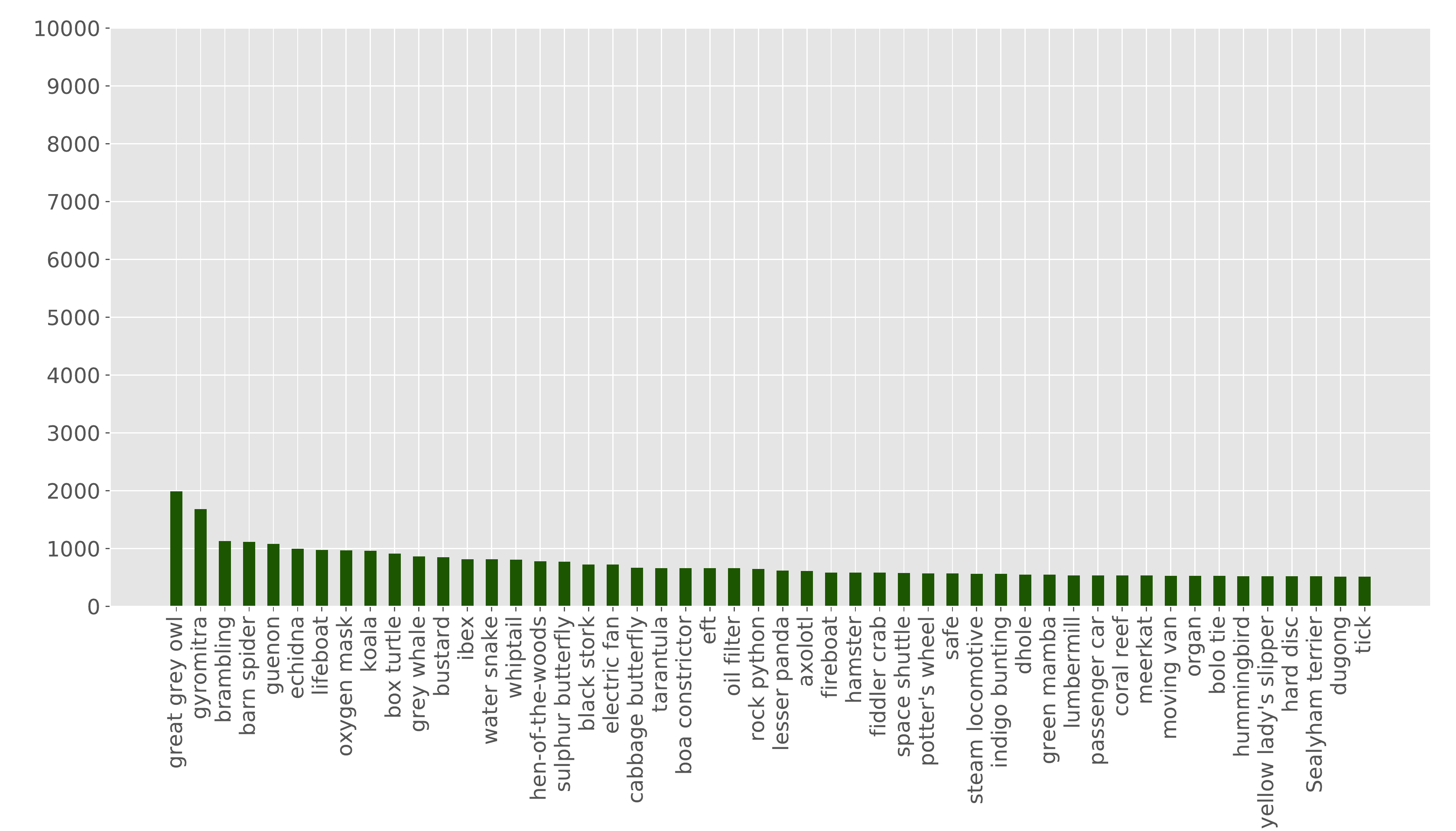}}
	\caption{
		We ran LIME and LIME-G on 200 images, each run has 500 intermediate perturbation samples.
		Here, for LIME (a) and LIME-G samples (b), we show a histogram of the top-1 predicted class labels for all $200$ runs $\times 500$ samples = 100,000 images.
		The set of 200 images comprises of cases where LIME-G \underline{outperformed} (100 images) and \underline{underperformed} (100 images) LIME on the SSIM sensitivity metric (Sec.~\ref{sec:sensitivity}).
		LIME perturbed samples are highly biased towards few \class{jigsaw~puzzle}, \class{maze} classes (top panel), which is somewhat intuitive given the gray-masked images (see Figs.~\ref{fig:LIME_ImageNet_top_sensitivity_1}--\ref{fig:LIME_ImageNet_bottom_sensitivity_2}). 
		In contrast, the histogram of LIME-G samples are almost uniform.
		\textbf{x-axis:} For visualization purposes, we sorted the top-1 labels and showed only first 50 labels.
	}
	\label{fig:LIME_histogram_full}
\end{figure}

\begin{figure}[h]
	\begin{tabular}{cc}
		\includegraphics[width=0.32\linewidth]{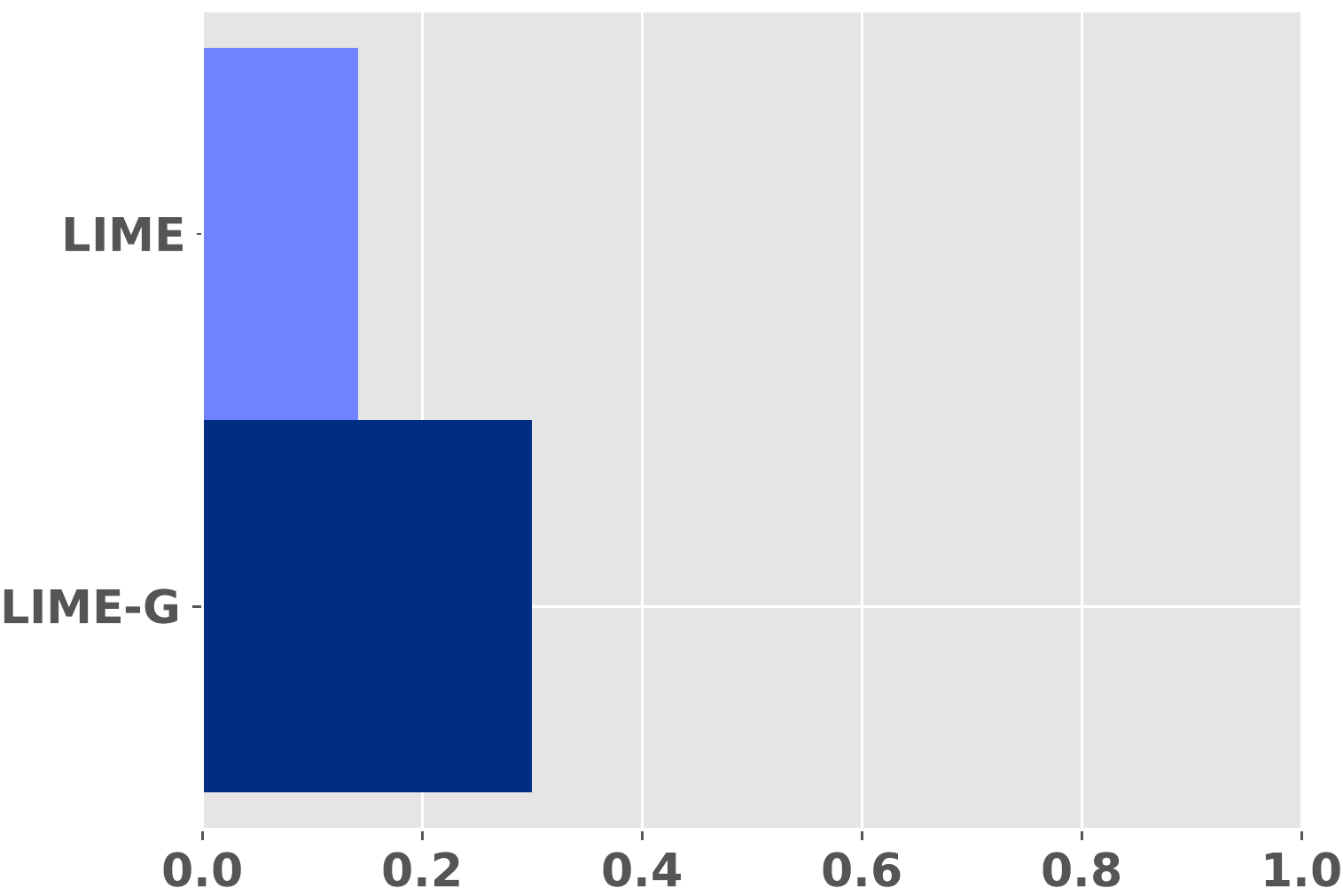} 
		\includegraphics[width=0.32\linewidth]{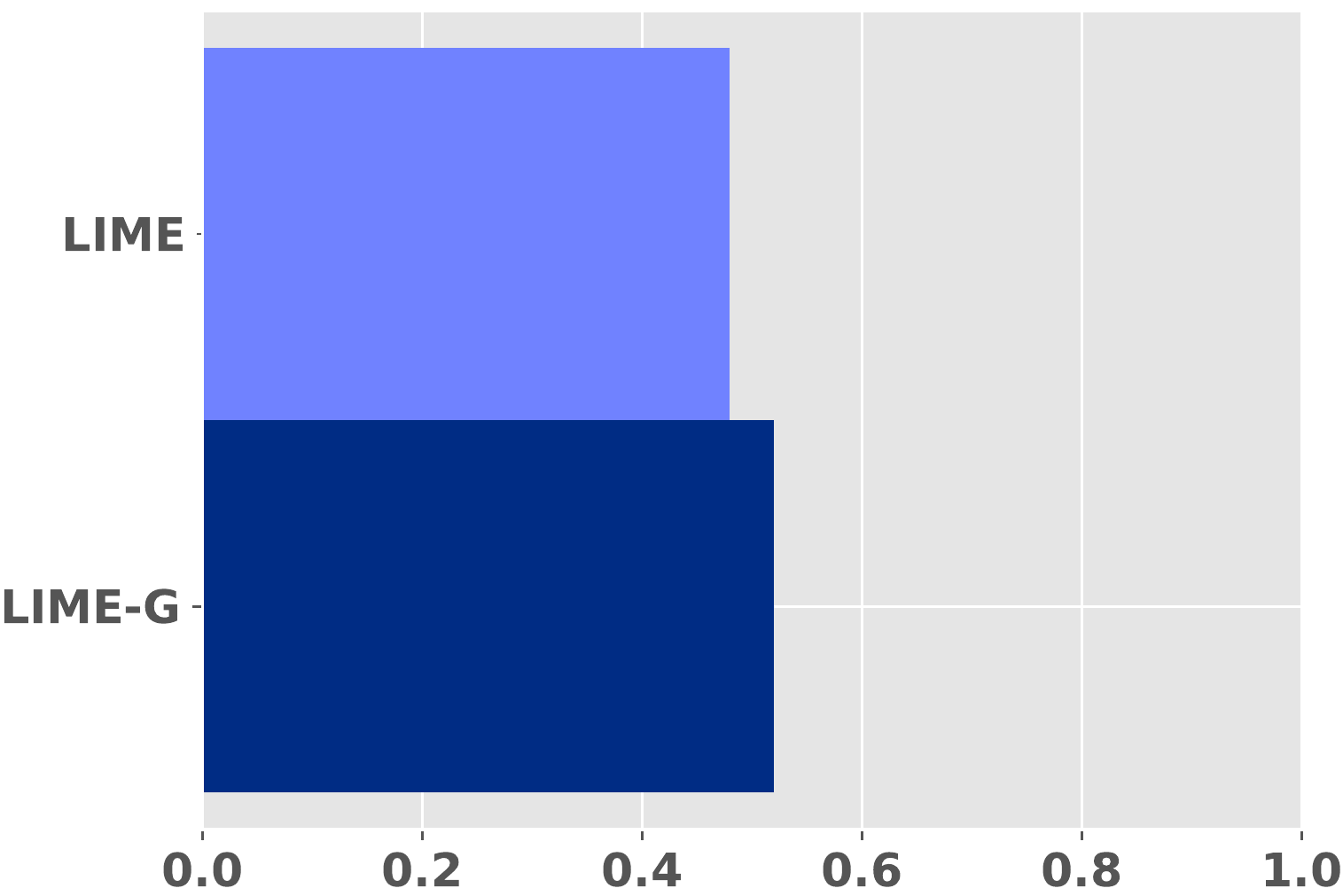}   
		\includegraphics[width=0.32\linewidth]{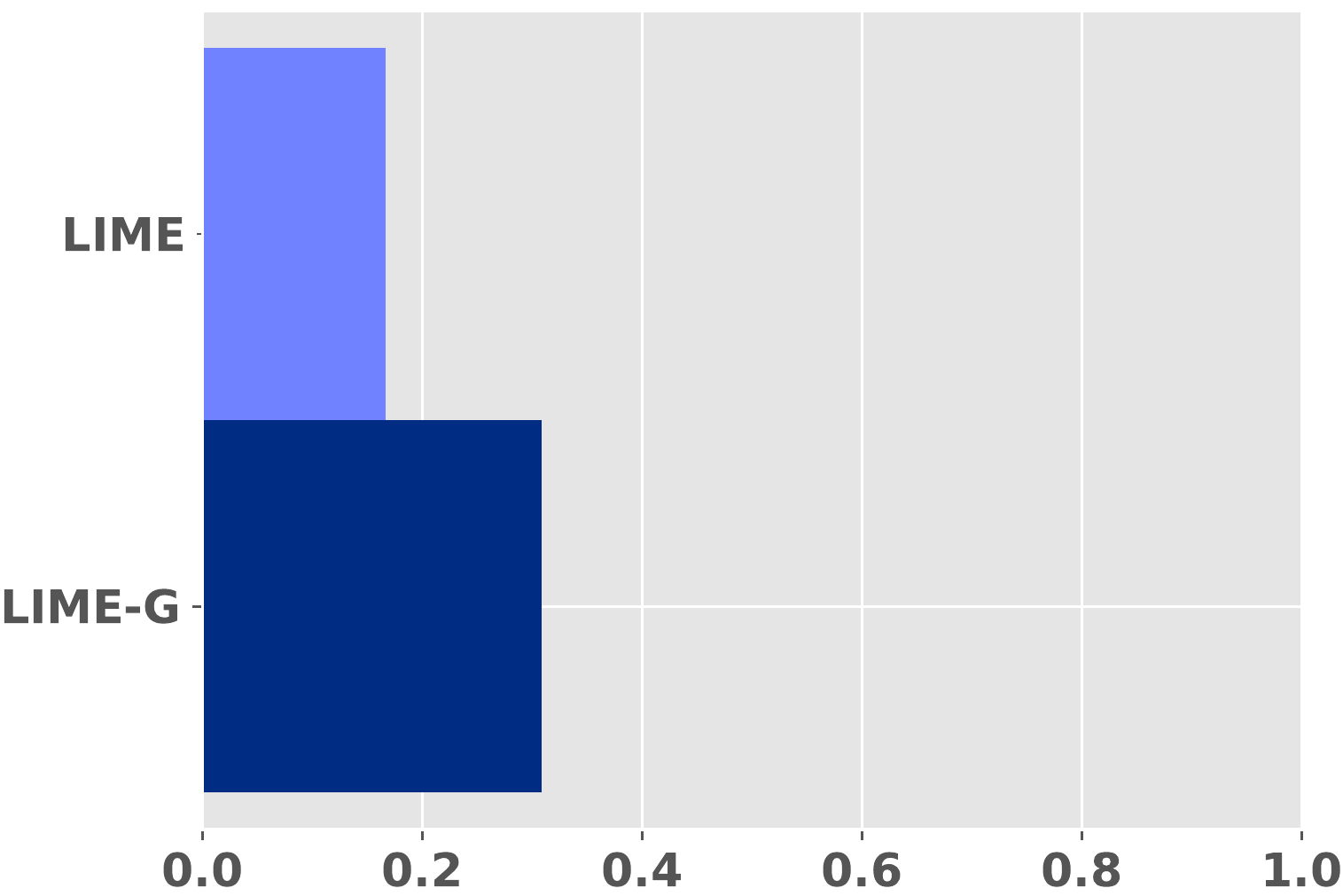}\\
		{	
			\hskip 0.6in (a) SSIM
			\hskip 0.2in (b) Pearson correlation of HOG features
			\hskip 0.02in (c) Spearman rank correlation
		}
	\end{tabular}
	\caption{
		Bar plots comparing the LIME vs. LIME-G robustness (higher is better) across two different numbers of superpixels $S \in \{50, 150\}$ under three different similarity metrics: SSIM (a), Pearson correlation of HOG features (b), 
		and Spearman rank correlation (c).
		For each image in 1000 random ImageNet-S images, we produced a pair of heatmaps by running LIME (light-blue) or LIME-G (dark-blue) with two different numbers of superpixels $S \in \{50, 150\}$.
		Each bar shows the mean similarity across all 1000 heatmap pairs.
		LIME-G is consistently more robust than LIME, specifically by $\sim$200\% under the SSIM (a) and Spearman rank correlation (c).
	}
	\label{fig:LIME_superpixel_sens}
\end{figure}

\end{document}

%% file: math_commands.tex
\usepackage{amsmath,amsfonts,bm}

\def\eqref#1{equation~\ref{#1}}

\def\1{\bm{1}}

\def\va{{\bm{a}}}

\def\vf{{\bm{f}}}

\def\vx{{\bm{x}}}

\def\mA{{\bm{A}}}

\DeclareMathAlphabet{\mathsfit}{\encodingdefault}{\sfdefault}{m}{sl}
\SetMathAlphabet{\mathsfit}{bold}{\encodingdefault}{\sfdefault}{bx}{n}

\def\sR{{\mathbb{R}}}

\newcommand{\E}{\mathbb{E}}

\DeclareMathOperator*{\argmin}{arg\,min}